\documentclass[acmtog]{acmart}

\usepackage{multirow}

\makeatletter
\AddToHook{begindocument/end}{%
  \ifnum\@abspage@last=1\relax
    \xdef\@abspage@last{\number\maxdimen}%
  \fi}
\makeatother

\begin{document}

\setcopyright{cc}
\setcctype{by}
\acmJournal{TOG}
\acmYear{2026} \acmVolume{45} \acmNumber{6} \acmArticle{264}
\acmMonth{12} \acmDOI{10.1145/3842587}

\title{AnyStitch: Generative Large Parallax Image Stitching}

\author{Zixuan Chen}
\orcid{0009-0006-8283-3664}
\affiliation{%
  \department{Multimedia Laboratory}
  \institution{The Chinese University of Hong Kong}
  \city{Hong Kong}
  \country{China}}
\email{zxchen3350@gmail.com}

\author{Lingen Li}
\orcid{0000-0002-1313-8717}
\affiliation{%
  \department{Multimedia Laboratory}
  \institution{The Chinese University of Hong Kong}
  \city{Hong Kong}
  \country{China}}
\email{lgli@link.cuhk.edu.hk}

\author{Junhao Zhuang}
\orcid{0000-0002-1642-0750}
\affiliation{%
  \institution{Tsinghua University}
  \city{Beijing}
  \country{China}}
\email{zhuangjh23@mails.tsinghua.edu.cn}

\author{Ruikang Li}
\orcid{0009-0003-2608-2673}
\affiliation{%
  \department{Multimedia Laboratory}
  \institution{The Chinese University of Hong Kong}
  \city{Hong Kong}
  \country{China}}
\email{1155231343@link.cuhk.edu.hk}

\author{Tianfan Xue}
\authornote{Corresponding author.}
\orcid{0000-0001-5031-6618}
\affiliation{%
  \department{Multimedia Laboratory}
  \institution{The Chinese University of Hong Kong, Shanghai AI Laboratory, CPII under InnoHK}
  \city{Hong Kong}
  \country{China}
  }
\email{tfxue@ie.cuhk.edu.hk}

\renewcommand\shortauthors{Chen et al.}

\begin{abstract}
Wide field-of-view panorama stitching is important for computational photography and immersive visual applications, but often faces some challenges like severe geometric disparities. The typical "warp-and-blend" stitching technique only works with limited parallax and fails on large-parallax scenes, due to its inherent restriction to 2D geometric transformations that cannot resolve underlying 3D structures. 
To break this limitation, we present AnyStitch, a generative framework that can merge multiple reference images with large parallax into a seamless panorama.
Unlike previous methods, we formulate the stitching problem as a reference-based image generation process, which naturally avoids stitching artifacts. Specifically, we propose a Hierarchical Projected Rotary Positional Encoding (HPRoPE) based on a 3D feed-forward model to establish accurate geometric relationships between the inputs and the panoramic canvas while preserving fidelity. 
Given that the resolution of the generated image varies with the number of inputs, we design a spatial autoregressive generation strategy to generate visually coherent panoramas at flexible resolutions. 
Finally, to enable the supervised fine-tuning of the proposed model, we construct a paired dataset utilizing 3D Gaussian Splatting (3DGS) scenes.
Experiments demonstrate that AnyStitch can seamlessly stitch a variable number of inputs (like 22 views in teaser) with large camera motion, support flexible control of target view and projection type, and handle simple dynamic scenes. Project page: \url{https://cecret3350.github.io/AnyStitch/}.
\end{abstract}

\ccsdesc[500]{Computing methodologies~Computational photography}

\keywords{Image stitching, computational photography, generative model}

\begin{teaserfigure}
\centering
  \includegraphics[width=1\textwidth]{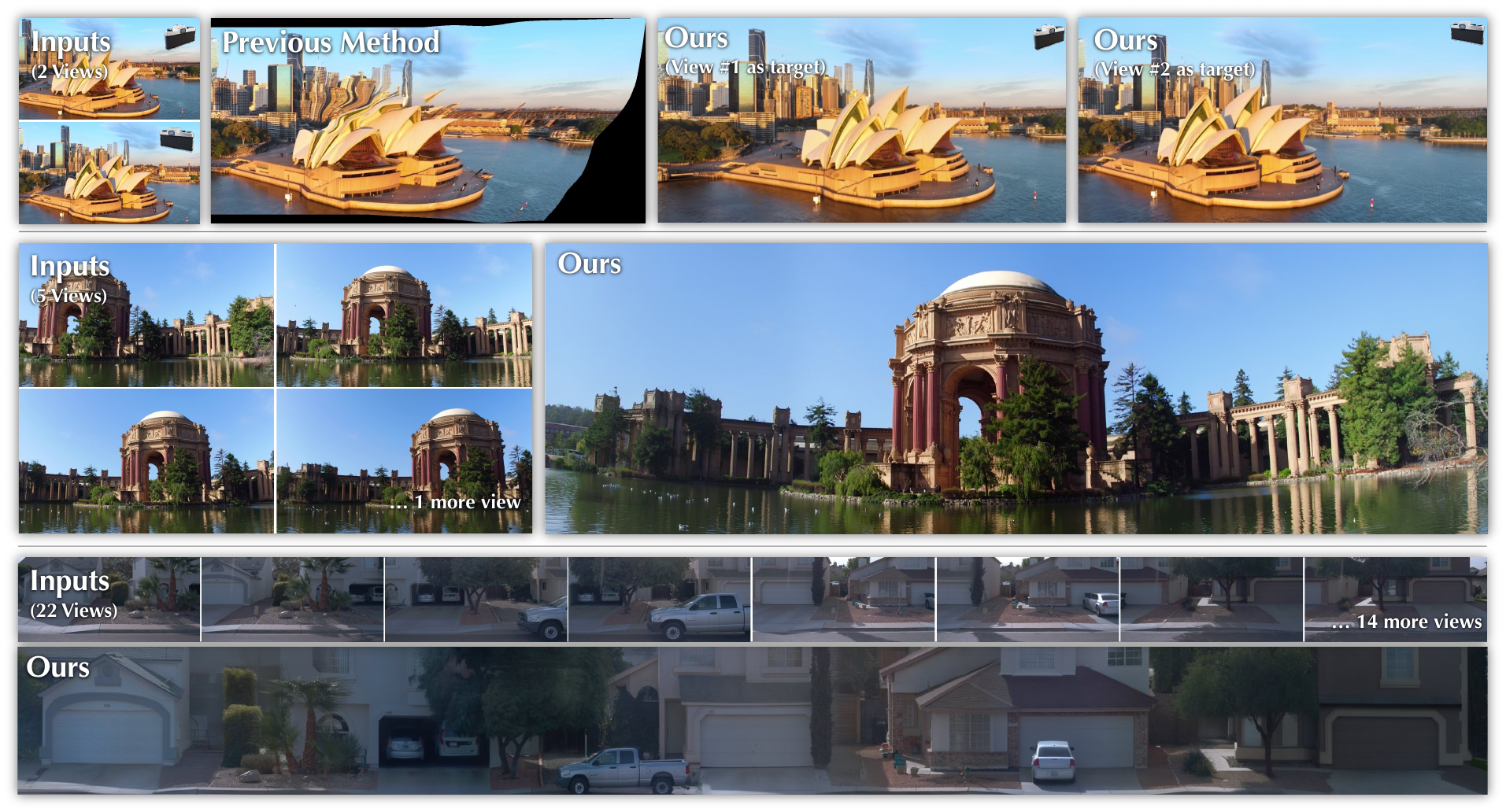}
  \caption{Panoramas synthesized by AnyStitch across three distinct scenes with varying projection types. \textbf{Top}: Drone aerial perspective (rectilinear projection). \textbf{Middle}: Casual walking-tour captures~\cite{Knapitsch2017} involving both camera translation and pure rotation (cylindrical projection). \textbf{Bottom}: Street-view sweeps~\cite{WaymoOpenDataset} featuring pure camera translation (parallel projection). AnyStitch seamlessly generates panoramas at flexible resolutions from both two-view and multi-view captures.}
  \label{fig:teaser}
\end{teaserfigure}

\maketitle

\section{Introduction}

How to capture an image with a wide field-of-view (FOV) that exceeds the camera capability has always been a fundamental problem in computer vision. 
To achieve this, many image stitching techniques are proposed to merge multiple inputs into a large seamless panorama. 
Typically, these methods follow a two-stage paradigm. They first warp the input images using homographies~\cite{AutoStitch, APAP, SPW, ELA, WRBIS, TFA} or optical flow~\cite{PixelStitch, PWANet}, and then fuse the aligned images to form the final panorama.
Despite all recent advances, these 2D warping paradigms are intrinsically limited to scenes with small parallax. Once large parallax occurs, even when there are still substantial overlapping regions between the input views, previous methods fail severely, as shown in the top scene of Fig.~\ref{fig:teaser}.

Due to this limitation, users should keep their cameras as stationary as possible, which restricts the flexibility of capture and the diversity of applicable scenarios.
Imagine a casual "walk-and-shoot" scenario around a large architectural landmark or along a long street, as shown in Fig.~\ref{fig:teaser}.
The continuous camera translation introduces massive parallax across multiple views. In such cases, previous methods are unable to disentangle the overlapping 3D structures.
Therefore, in this work, we study the following question: \textit{can we seamlessly merge multiple inputs with a large parallax to create a coherent visual representation?}

Inspired by the remarkable capability of state-of-the-art image generation models to synthesize coherent visual content, we answer this question from a generative perspective. Unlike traditional image stitching methods, we formulate the stitching problem as a reference-based image generation and propose a novel framework, \textbf{AnyStitch}. In our design, the final stitched image is generated by taking multiple large-parallax captures as references. Because the output is generated rather than simply blended, this approach inherently circumvents the persistent seams and ghosting artifacts that plague traditional methods.

However, directly applying image generation models to image stitching by taking multiple inputs as conditions is far from a trivial task due to the following reasons. First, existing generative models typically assume that the reference latent shares the same spatial positional encoding as the noisy latent. 
By contrast, in image stitching, the output resolution is substantially larger than the input, often scaling up by 5 to 10 times with a dramatically altered aspect ratio (\textit{e.g.}, from a standard 2:1 input to a 10:1 panoramic output, as shown in Fig.~\ref{fig:teaser}). As a result, the input condition occupies only a small, localized region within the expansive final canvas.
Relying on standard positional encodings forces the model to depend exclusively on attention mechanisms to infer spatial correspondences. This compromises geometric accuracy and leads to noticeable stitching errors, as shown in Fig.~\ref{fig:ab_rope_syn}. In addition, it is computationally infeasible for generation models to generate the entire stitched image in a single pass~\cite{CubeComposer}. Second, fine-tuning a generative model is inherently a supervised process, whereas previous learning-based stitching methods depend on unsupervised training. There is a critical absence of a dataset containing large-parallax inputs paired with geometrically consistent ground truth.

To address these challenges, we design AnyStitch, a novel stitching model combines the geometric perception with generative model.
First, to deal with large parallax, we move beyond the traditional 2D warping paradigm to 3D modeling. While 2D-based methods typically collapse under large parallax due to their restrictive assumption of a planar scene, a 3D feed-forward model~\cite{DUST3R,VGGT, depthanything3,Pi3} enables more robust 3D geometric grounding~\cite{LiftProj, PIS3R}. This shift allows the model to perceive the underlying 3D structure of the scene, effectively disentangling overlapping foreground and background elements that are otherwise inseparable in 2D pixel space. Specifically, we utilize 3D model to estimate depth and camera pose of each input. Then, we can flexibly define a target view and projection type. With these geometric priors, we are able to determine the resolution of the final canvas and establish the precise spatial mapping of each input onto it. 

To ensure this geometric information is properly used in a generative stitching model, we further introduce specialized mechanisms to generate the final panorama. A simple solution is to perform a hard projection of inputs in the pixel space, but it may suffer from severe structure distortions and information loss when depth estimation are inaccurate. Instead, we formulate the spatial mapping as a geometric prior. Specifically, we propose a Hierarchical Projected Rotary Positional Encoding (HPRoPE) that unifies the positional encodings of the input references and the generated image. Within HPRoPE, we allocate varying projection precisions across different attention heads to simultaneously ensure strict geometric accuracy and preserve content fidelity. As shown in Fig.~\ref{fig:ab_rope_syn} and Fig.~\ref{fig:ab_rope}, this prevents the structural distortions inherent to pixel-space projection. Second, to handle variable-resolution outputs and circumvent computational constraints, we introduce a patch-wise spatial autoregressive generation strategy. We generate the stitched panorama patch-by-patch using a sliding window, utilizing previously generated tokens as memory context to ensure consistency. Additionally, a relative positional encoding is integrated during this autoregressive process to prevent extrapolation artifacts. Finally, we bridge the data gap by constructing a synthetic dataset called AnyStitchSyn, utilizing 3D Gaussian Splatting (3DGS) scenes, rendering large-parallax inputs paired with precise ground truths to enable robust supervised fine-tuning.

Fig.~\ref{fig:teaser} demonstrates that AnyStitch can seamlessly integrate a variable number of large-parallax captures (like 20+ views shown in bottom) while offering flexible, user-defined control over target views and projection types, ultimately generating a visually coherent panorama at flexible resolutions.

\begin{figure*}[!t]
	\centering
	\includegraphics[width=\linewidth]{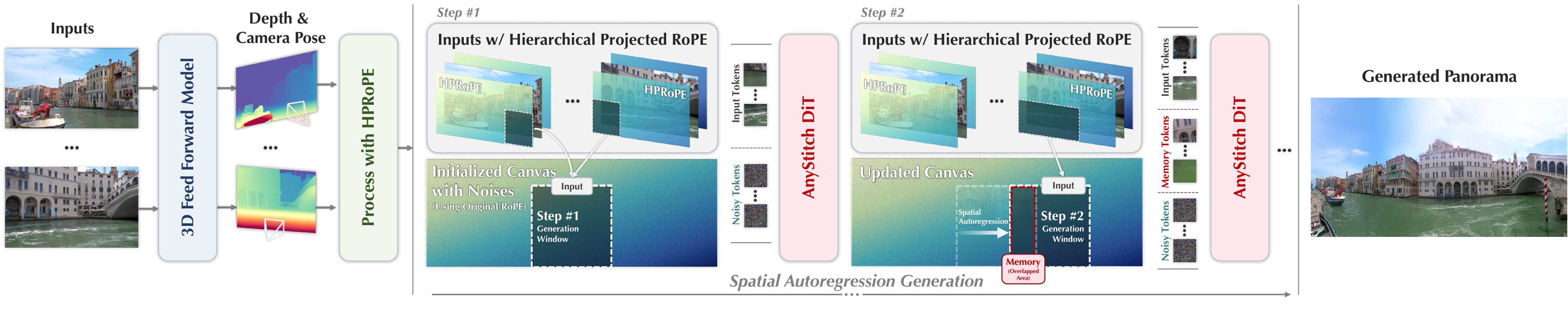}
	\caption{The pipeline overview of AnyStitch. We first extract depths and camera poses via a 3D feed-forward model to determine the final size of panoramic canvas and compute the Hierarchical Projected RoPE (HPRoPE), establishing precise geometric conditioning. Subsequently, a Diffusion Transformer (DiT) synthesizes the panorama using a spatial autoregressive generation strategy. By processing the canvas patch-by-patch and incorporating previously decoded regions as memory tokens, our framework seamlessly merges large-parallax inputs into a visually coherent panorama at flexible resolutions.}
    \label{fig:pipeline}
\end{figure*}

\section{Related Work}

\paragraph{Image stitching}

Traditional image stitching frameworks rely on geometric features to establish correspondences, evolving from basic global homography alignments~\cite{AutoStitch} to more flexible spatially-varying warps~\cite{APAP, SPW, ELA, WRBIS, TFA}. To further mitigate the unnatural distortions introduced by these flexible deformations, subsequent approaches incorporated structure-preserving constraints, such as global similarity priors~\cite{SPHP, GSP, GES-GSP} and line-point consistency~\cite{LPC, DFW}, to maintain the integrity of salient geometric structures. 

With the success of Convolutional Neural Networks (CNNs) in other computer vision tasks, learning-based methods have emerged to extract high-level semantic features via neural networks. Recent learning-based paradigms have progressed from directly regressing global projective parameters~\cite{UDIS} to predicting sophisticated local mesh deformations~\cite{UDIS++, RopStitch} and dense bidirectional optical flow fields~\cite{PWANet, PixelStitch}. More recently, some methods have introduced generative models for image stitching. SRStitcher~\cite{SRStitcher} uses a pre-trained generative model to refine coarse fusion and rectangling results obtained from 2D-aligned inputs. GPIS~\cite{GPIS} further fine-tunes a generative model for each scene using 2D pre-aligned inputs and generates the panorama patch by patch with global canvas coordinates. Nevertheless, these methods either manipulate images through 2D pixel-space warping or rely on 2D pre-alignment, limiting their robustness under large parallax.

Beyond 2D warping, geometry-based methods construct panoramas through 3D reconstruction. Instant 3D Photography~\cite{Instant3DPhotography} uses captured depth to align RGB-D frames in 3D and fuse them into a panorama that supports parallax-aware rendering. More recently, feed-forward 3D reconstruction models (\textit{e.g.}, DUSt3R~\cite{DUST3R}, VGGT~\cite{VGGT}, Depth Anything 3~\cite{depthanything3}, and Pi3~\cite{Pi3}) have demonstrated efficient and generalizable reconstruction even from sparse image inputs. LiftProj~\cite{LiftProj} and PIS3R~\cite{PIS3R} build on these models~\cite{DUST3R, VGGT} to lift the inputs into 3D before projecting them onto the target view. These approaches improve geometric alignment, but their pixel-space rendering remains sensitive to errors in the estimated geometry.

\paragraph{Generative novel view synthesis}
Generative novel-view synthesis combines learned priors with geometric guidance to synthesize content beyond the observed views. Context-aware layered depth inpainting~\cite{3DPhotoInpainting} synthesizes color and depth in disoccluded regions for novel-view rendering. GenWarp~\cite{GenWarp} augments generative novel-view synthesis with geometric warping signals. ViewCrafter~\cite{ViewCrafter} conditions video diffusion on point-based 3D cues. GEN3C~\cite{GEN3C} renders a 3D cache along a target camera trajectory to guide video generation. These methods primarily target novel-view image or video synthesis, whereas AnyStitch integrates multiple casually captured inputs into a single panoramic canvas.

\paragraph{Geometric positional encoding}
Geometric positional encodings incorporate camera and projection geometry into token positions, enabling attention to reason about spatial correspondences across views. PRoPE~\cite{PRoPE} encodes relative projective transformations between complete camera frustums, incorporating both camera intrinsics and extrinsics into multi-view attention. RayRoPE~\cite{RayRoPE} associates each image patch with a camera ray and predicts a point along the ray, whose query-frame projection defines a geometry-adaptive positional encoding for multi-view attention. PE-Field~\cite{PEField} and MosaicMem~\cite{MosaicMem} further introduce geometry-aware positional encodings into generative models. They both establish source-to-target correspondence without explicitly warping the source RGB content, instead assigning RoPE to source tokens according to their reprojected coordinates in the target view. Unlike these methods, HPRoPE is specifically tailored to multi-image stitching with a dual-precision projection design, where separate attention heads assign input tokens to panorama coordinates using dense-depth and median-depth projections, respectively.

\section{Methodology}

Given a variable number of casually captured input images $\{I_i\}_{i=1}^N$, a user-specified target view $V_{\textit{target}}$ represented by intrinsic and extrinsic matrix $(\mathbf{K}_{tgt}, \mathbf{T}_{tgt})$, and a user-specified projection type $\mathcal{P}$ (\textit{e.g.}, rectilinear, cylindrical, and equirectangular projection), our goal is to stitch these inputs into a seamless panorama $I_{out}$. Previous methods typically warp and blend the inputs, making them highly sensitive to large parallax. 
In contrast, to address the challenging alignment of input images under motion parallax, we use a feed-forward 3D reconstruction model to estimate the camera pose and depth map for each input image. Using the estimated geometry, we map each input patch to its corresponding location on the output canvas and encode this correspondence by modifying its RoPE, without explicitly warping the input content. Then we finetune the pretrained generative model with these virtually warped patches as condition, to ensure they can be naturally stitched together.

In addition, we also devise a spatial autoregressive generation strategy including memory mechanisms and relative positional encoding to enable large panorama generation with a variable number of inputs. The details of each module are described below.

\begin{figure}[!t]
	\centering
	\includegraphics[width=\linewidth]{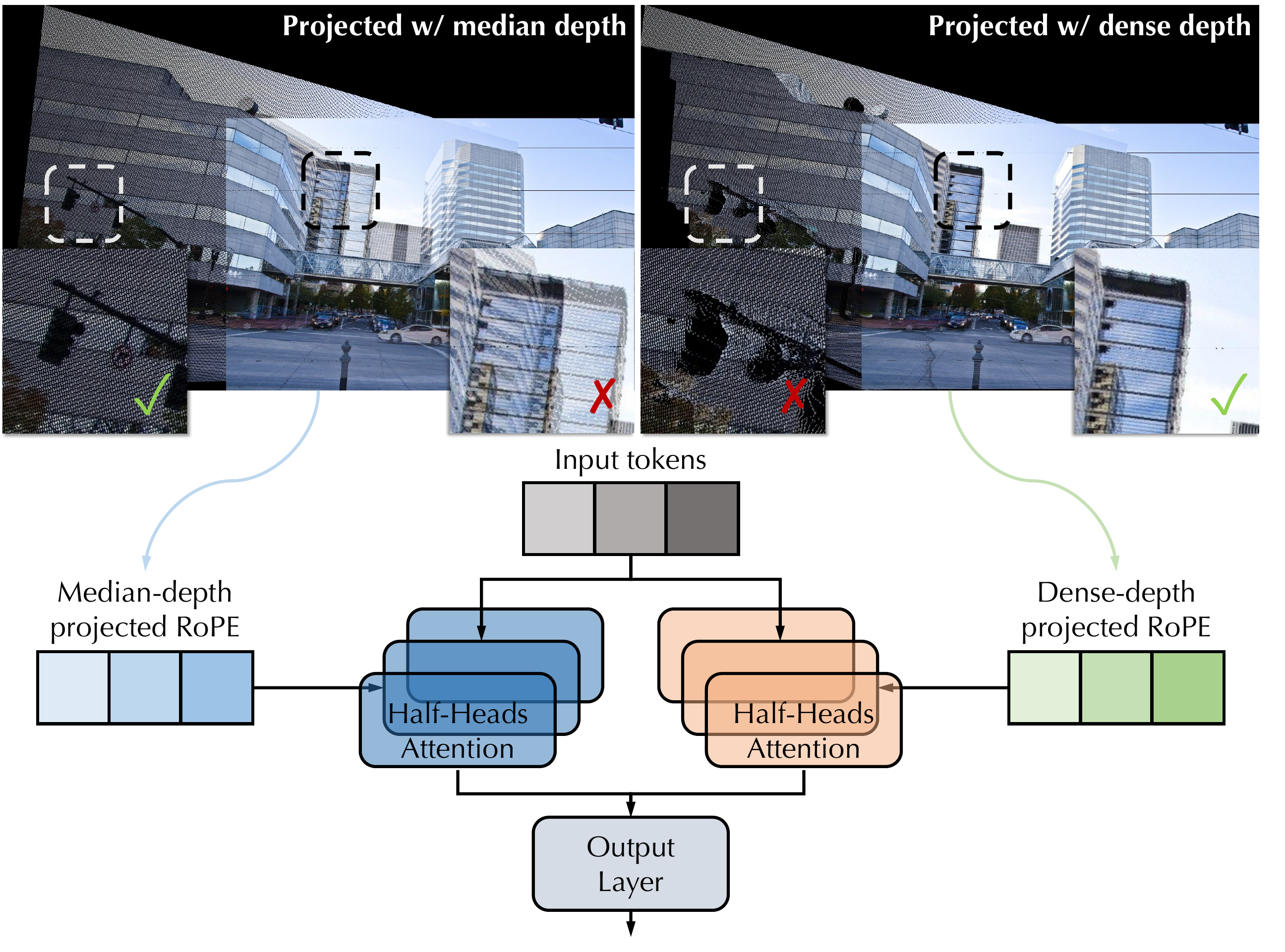}
	\caption{Illustration of HPRoPE. Dense-depth projection captures fine-grained geometry but can lose information around inaccurate depth boundaries, while median-depth projection provides coarse yet stable alignment that better preserves content fidelity. HPRoPE evenly assigns the two projected RoPE variants to separate halves of the attention heads and the output layer is trained to smartly extract and combine their outputs.}
    \label{fig:2head_rope}
\end{figure}

\subsection{Hierarchical Projected RoPE}
Standard diffusion transformers \cite{DiT} rely on 2D RoPE to model spatial relationships. This standard RoPE fundamentally assumes that the reference conditions and the generated latent represent the same scene layout and share an identical, uniform spatial grid. However, image stitching fundamentally violates this assumption. In large-parallax stitching scenarios, casually captured inputs $\{I_i\}_{i=1}^N$ arbitrarily occupy localized, unaligned regions within a substantially larger panoramic canvas. Standard 2D RoPE fails to establish the correct spatial coordinates for input references, leading to severe spatial misalignment during the generative process.

To establish accurate spatial correspondences, we leverage a 3D feed-forward model. By explicitly modeling 3D geometry through depth and camera poses, it naturally overcomes the inherent limitations of traditional 2D transformations (\textit{e.g.}, homography or optical flow) in large-parallax scenes, providing accurate geometric priors \cite{LiftProj, PIS3R}.

Specifically, we first employ Depth Anything 3 (DA3)~\cite{depthanything3} to estimate the intrinsic matrix $\mathbf{K}_i$, extrinsic matrix $\mathbf{T}_i$ and a corresponding dense depth map $D_i$. With these estimations, we can project each original spatial RoPE $(u_i, v_i)$ from the input image plane into a unified 3D space, and then project to the panorama canvas via the given target view $V_{target}$ and the projection type $\mathcal{P}$ \cite{MosaicMem}:
\begin{equation}
    (u'_i, v'_i) = \Pi_{\mathcal{P}} \left( \mathbf{K}_{tgt} \mathbf{T}_{tgt} \mathbf{T}_i^{-1} \mathbf{K}_i^{-1} (u_i, v_i, D_i) \right),
    \label{eq:eq1}
\end{equation}
where $\Pi_{\mathcal{P}}(\cdot)$ denotes the project operation of $\mathcal{P}$.

While Eq. \ref{eq:eq1} provides a rigorous geometric alignment, it heavily relies on the accuracy of the estimated dense depth map $D_i$. In casually captured large-parallax scenes, depth estimation inevitably suffers from local errors, particularly around occlusion boundaries and fine structures. Strictly enforcing such fine-grained but noisy 3D priors across all attention layers can overly constrain the generative process, occasionally producing local distortions or spatial offsets in the generated panorama.

To mitigate this issue, we propose Hierarchical Projected RoPE (HPRoPE). As shown in Fig.~\ref{fig:2head_rope}, instead of relying exclusively on the dense depth $D_i$, we introduce a robust auxiliary planar projection driven by the scalar median depth of the input view, denoted as $\bar{D}_i = \text{median}(D_i)$. This alternative spatial mapping is formulated as:

\begin{equation}
(\bar{u}'_i, \bar{v}'_i) = \Pi_{\mathcal{P}} \left( \mathbf{K}_{tgt} \mathbf{T}_{tgt} \mathbf{T}_i^{-1} \mathbf{K}_i^{-1} (u_i, v_i, \bar{D}_i) \right).
\label{eq:eq2}
\end{equation}

This dual-branch design offers obvious advantages, as shown in Fig.~\ref{fig:2head_rope}. The dense-depth projection (Eq.~\ref{eq:eq1}) provides an accurate geometric layout, while the median-depth projection (Eq.~\ref{eq:eq2}) treats the scene as a simple 3D plane, providing a stable baseline that is unaffected by local depth noise.

Finally, we downscale the coordinates and obtain $(u_{l, i}, v_{l, i}) = (u'_i/s, v'_i/s)$ and $(\bar{u}_{l, i}, \bar{v}_{l,i}) = (\bar{u}'_i/s, \bar{v}'_i/s)$ for latent space, where $s$ denotes the downsampling factor from image space to latent space. By integrating the input index $t$ and the text index $c$, we construct two sets of RoPE coordinates for each latent token: the dense-depth coordinates $(t_i, u_{l, i}, v_{l, i}, c_i)$ and the planar median-depth coordinates $(t_i, \bar{u}_{l, i}, \bar{v}_{l, i}, c_i)$. To empower the self-attention layer to process these dual geometric representations concurrently, we introduce a head-wise coordinate assignment mechanism. Given an attention layer with $H$ heads, we partition them into two disjoint subsets: $\mathcal{H}_{dense}$ and $\mathcal{H}_{median}$. The heads are divided evenly between the two subsets, such that $|\mathcal{H}_{dense}| = |\mathcal{H}_{median}| = H/2$. For a specific head $h$, the applied rotary positional encoding is explicitly conditioned on its designated subset:
\begin{equation}
\text{RoPE}_h(\mathbf{x}_{h,i}) =
\begin{cases} 
\mathcal{F}_{rot}(\mathbf{x}_{h,i} \,|\, t_i, u_{l, i}, v_{l, i}, c_i), & \text{if } h \in \mathcal{H}_{dense} \\
\mathcal{F}_{rot}(\mathbf{x}_{h,i} \,|\, t_i, \bar{u}_{l, i}, \bar{v}_{l, i}, c_i), & \text{if } h \in \mathcal{H}_{median} 
\end{cases}
\end{equation}
where $\mathbf{x}_{h,i}$ represents the query or key feature of token $i$ in the $h$-th head, and $\mathcal{F}_{rot}(\cdot)$ denotes the standard rotary transformation function utilizing the given base coordinates. By assigning these complementary coordinates to separate attention heads, HPRoPE enables each attention layer to jointly exploit fine-grained geometric alignment and a robust global layout. Moreover, it introduces no additional parameters and negligible computational overhead, since it only applies precomputed coordinates to existing attention heads.

\subsection{Spatial Autoregressive Generation}
Unlike most image processing and generation tasks, the output size of image stitching is decided by the number and spatial arrangement of captured inputs.
While state-of-the-art DiTs support variable sizes, their generation quality degrades severely when extrapolated beyond a limited resolution range (most modern DiTs struggle to generate output larger than 4K, which is even not sufficient for image stitching). 
Instead of attempting to denoise this massive panorama in a single computationally prohibitive and unstable pass, we progressively generate content patch by patch on the final canvas.

\begin{figure}[!t]
	\centering
	\includegraphics[width=\linewidth]{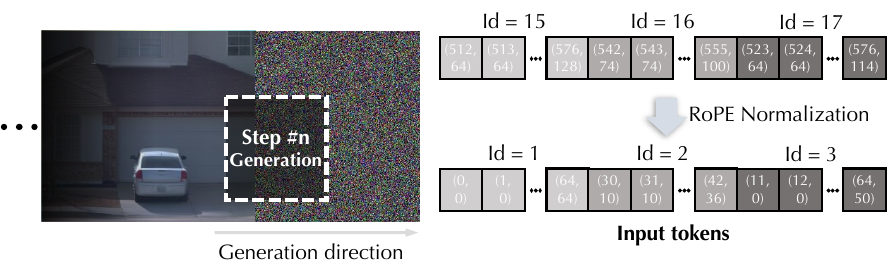}
	\caption{Relative RoPE in each step of spatial autoregressive generation. 
    }
    \label{fig:relative_rope}
\end{figure}

First, after projecting all input views using the specified target view and projection type, we define the spatial extent of the target canvas as the minimum bounding rectangle enclosing all valid dense-depth projected coordinates $\{(u'_i, v'_i)\}$. This bounding rectangle determines the field of view and resolution of the output panorama. Based on this size, we initialize the panorama as pure Gaussian noise.
Specifically, as shown in Fig.~\ref{fig:pipeline}, we start the generation from the spatial center of the panorama, processing local patches of size $L \times L$ and progressively expanding outwards with a stride of $S$. The traversal proceeds column by column from the central column, while patches within each column are generated from the center toward the bottom and top.
In each patch-wise generation, the input to our AnyStitch DiT is constructed by concatenating three distinct types of tokens: 1) noisy tokens cropped from the pure noisy panorama in the current $L \times L$ spatial window, 2) input tokens which are routed into the current generation window according to whether its dense-depth projected coordinate $(u'_i, v'_i)$ falls within the patch,
3) memory tokens that sampled from the overlapping regions of the previously generated panoramic patches, ensuring consistent generation between adjacent patches (see Fig.~\ref{fig:ab_ar_consistency}).

Moreover, we observe that DiTs' extrapolation failure in large panorama generation typically manifests in two-folds: spatial dimension and sequential dimension. On the one hand, when the panoramic canvas size significantly exceeds the resolution encountered during training, the absolute spatial frequencies in RoPE shift out of the stable distribution, leading to structural collapse. On the other hand, as the number of input views increases, the sequential indices $t_i$ assigned to input tokens similarly encounter extrapolation issues.
To address these challenges, we introduce a relative positional encoding mechanism, as illustrated in Fig.~\ref{fig:relative_rope}. Specifically, within each local generation context, we jointly shift both branches of spatial RoPE by the shared minimum coordinate along each spatial axis and shift the input sequence indices by their minimum value.
This operation ensures that both spatial and sequential frequencies consistently remain within a stable, pre-trained range, regardless of the global canvas size or the total number of input images. By circumventing these two types of extrapolation failures, AnyStitch achieves robust panorama generation on a large canvas.

\subsection{AnyStitchSyn Dataset}
Most existing learning-based image stitching frameworks follow UDIS's unsupervised paradigms~\cite{UDIS}, primarily focusing on feature matching and warping. However, this approach inherently conflicts with the DiTs, which fundamentally depend on high-fidelity ground truths to learn the distribution. Furthermore, the image stitching community has long suffered from a critical absence of training datasets that provide casually captured, large-parallax inputs paired with geometrically consistent panoramic grount truths. 
To bridge this gap, we introduce the AnyStitchSyn Dataset, a synthetic dataset tailored for large-parallax stitching.

\begin{figure}[!t]
	\centering
	\includegraphics[width=\linewidth]{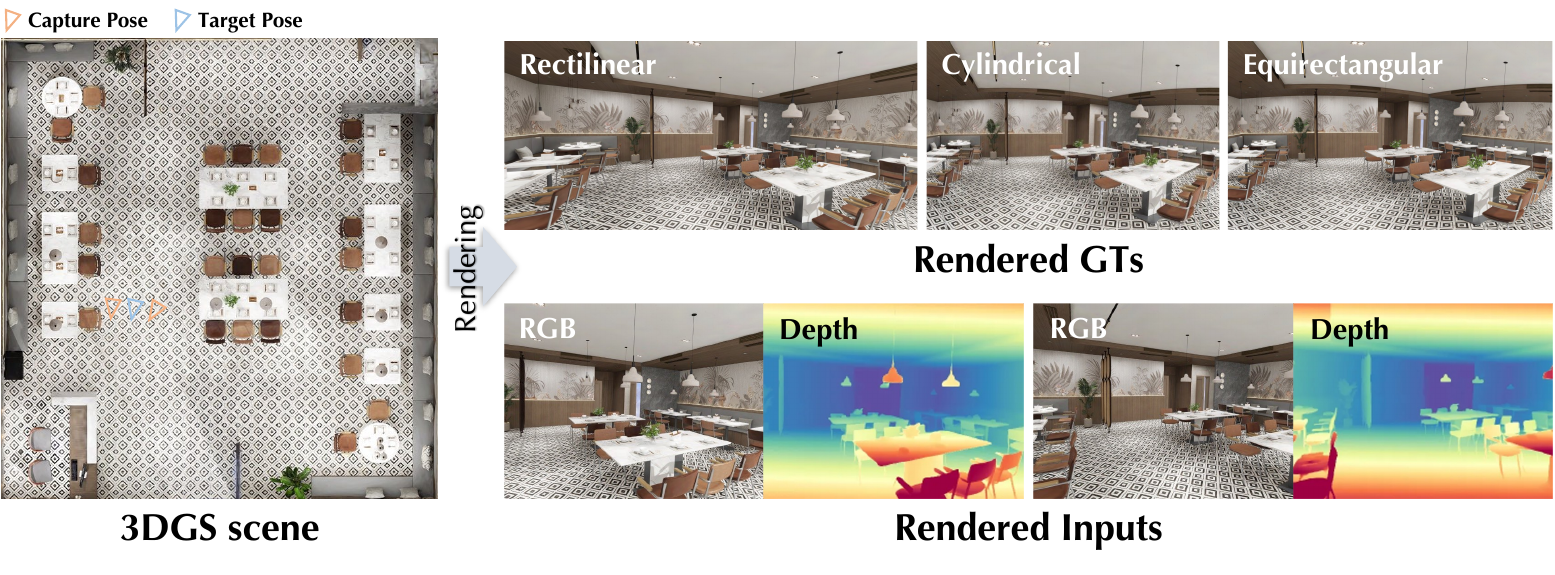}
	\caption{Rendering examples of our synthetic dataset.}
    \label{fig:gensyndata}
\end{figure}

Leveraging the photorealistic rendering capabilities and explicit geometric control of 3D Gaussian Splatting (3DGS)~\cite{3DGS}, we can synthesize inputs from views we want, introducing large disparities. Moreover, we can also synthesize perfectly aligned ground truths for corresponding inputs. We visualize the capturing process within 3DGS scenes in Fig.~\ref{fig:gensyndata}. We first randomly sample a camera pose to serve as the initial capture view. To simulate natural hand-held camera movements and induce parallax, we translate the camera by a random distance and inject rotational perturbations to obtain the second capture view. The corresponding RGB images and dense depth maps for both input views are then rendered. 

\begin{figure*}[!t]
	\centering
	\includegraphics[width=\linewidth]{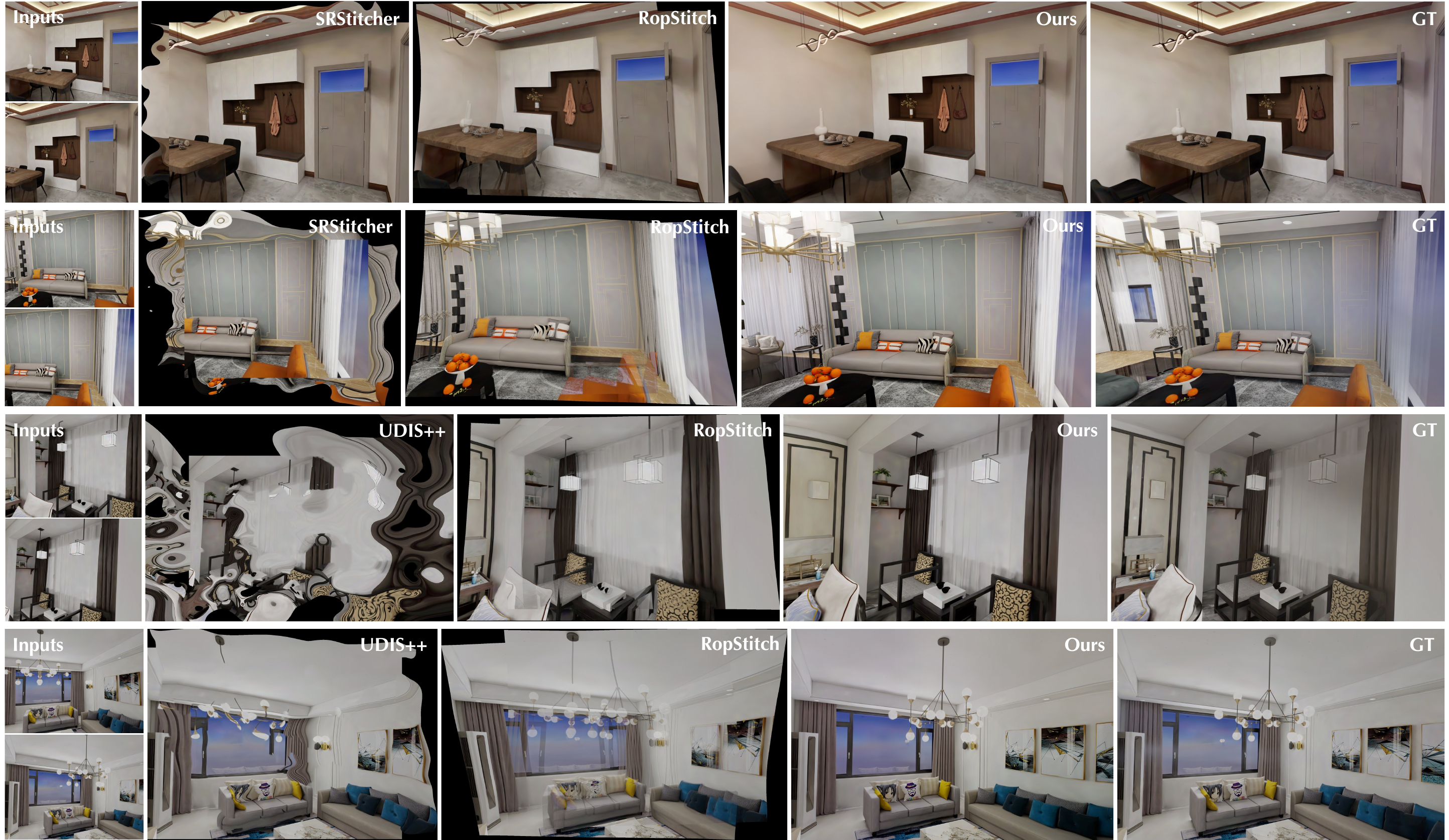}
	\caption{Visual comparisons on AnyStitchSyn test set. Our method uses DA3-estimated geometry.}
    \label{fig:comparison_anystitchsyn}
\end{figure*}

\begin{table}[t]
    
	\centering
	\caption{Full-reference comparison on the AnyStitchSyn test set. AnyStitch uses camera poses and depth maps estimated by DA3 rather than the ground-truth camera poses and rendered depth maps available from the 3DGS scenes. For each method, metrics are computed over the non-black region shared by its output and the ground truth.}
	\label{tab:full_reference}
    \begin{tabular}{c|c|c}
        \toprule[1.2pt]
        \multirow{2}{*}{Method} & \multicolumn{2}{c}{AnyStitchSyn$^{\mathrm{w/\ DA3\ geometry}}$} \\
        & SSIM $\uparrow$ & LPIPS $\downarrow$ \\
        \midrule \midrule
        UDIS++~\cite{UDIS++} & 0.7471 & 0.5255 \\
        SRStitcher~\cite{SRStitcher} & 0.7165 & 0.5408 \\
        RopStitch~\cite{RopStitch} & 0.7487 & 0.5052 \\
        \midrule
        AnyStitch & \textbf{0.8061} & \textbf{0.3200} \\
        \bottomrule[1.2pt]
    \end{tabular}
\end{table}

To construct the panoramic ground truth, we define the target camera pose as the interpolated average of the input views. 
Three distinct types of panoramic ground truths from this target pose are rendered, \textit{i.e.}, rectilinear, cylindrical and equirectangular.
For the rectilinear ground truth, we directly render the scene using standard perspective projection with an enlarged field-of-view (FOV). For the cylindrical and equirectangular ground truths, we first construct an intermediate environmental cubemap by rendering six orthogonal perspective images from the target pose. Next, we formulate dense 3D viewing rays based on continuous cylindrical and spherical (longitude-latitude) coordinates, respectively. By projecting these specific rays onto the cubemap faces and performing bilinear sampling, we effectively obtain seamless cylindrical and equirectangular panoramas.

Last, we utilize Eq.~\ref{eq:eq1} to explicitly project the input views onto the chosen ground truth canvas. By computing the minimum bounding rectangle that tightly encloses these projected coordinates, we crop the full panorama to yield the final, perfectly aligned ground truth tailored for the given inputs. 

We synthesize our dataset using 1000 scenes from InteriorGS dataset\cite{InteriorGS}. We utilize the first 950 scenes for training, randomly generating 1 to 3 data groups per scene to collect a total of 2,192 training pairs. Crucially, each training pair includes two input views alongside corresponding ground-truth panoramas rendered in three different projection models (The effectiveness analysis of this mixed-projection data is provided in supplementary material). For evaluation, we extract 100 test pairs from the remaining 50 scenes and adopt rectilinear projection for ground truth.

\section{Experiment}

\subsection{Experimental Setting}
\paragraph{Implementation details}
We implement AnyStitch built upon \\ FLUX.2-klein-9B~\cite{FLUX2}, a state-of-the-art Diffusion Transformer (DiT) designed for image generation and editing. To adapt the powerful pre-trained generative prior encapsulated in this model to the large-parallax image stitching task, we fine-tune the network on our AnyStitchSyn Dataset.
Specifically, we employ Low-Rank Adaptation (LoRA)~\cite{LoRA} and set LoRA rank to 32. The model is fine-tuned for 6 epochs with a batch size of 4. We optimize the network using the Adam optimizer with a constant learning rate set to $1 \times 10^{-4}$. Training takes approximately 8 hours on 4 NVIDIA A800 equivalent GPUs. At inference, we use a patch size $L$ of 1024 and a stride $S$ of 768, corresponding to a 25\% overlap, together with 40 denoising steps. More training details are available in the supplementary.

\paragraph{Evaluation metrics}
Beyond conventional full-reference metrics commonly used in image processing (\textit{e.g.}, PSNR, SSIM~\cite{SSIM} and LPIPS~\cite{LPIPS}), evaluating large-parallax image stitching also requires assessing semantic and perceptual coherence.
To this end, we employ two reference-free metrics: the Stitching Image Quality Score (SIQS) and Multi-Image Comparative Quality Score (MICQS)~\cite{SIQS}. SIQS leverages the advanced reasoning capabilities of large multimodal models~\cite{Qwen-VL} to grade the stitched panoramas across five critical perceptual dimensions: seam visibility, brightness transition, geometric distortion, clarity, and the presence of abnormal content. Meanwhile, MICQS utilizes large multimodal models as an impartial judge to evaluate and compare the stitched panoramas based on the aforementioned five perceptual dimensions.
In addition, we also conduct human subjective evaluation. Specifically, for each dataset, we randomly select 20 scenes and invite 51 users to participate. During the evaluation, users are asked to perform pairwise comparisons between the stitching results of our model and a randomly selected baseline method.

\begin{table}[t]
	\centering
	\caption{Quantitative comparison on two large-parallax datasets. $^\dagger$ denotes our reproduction of a PIS3R/LiftProj-style pipeline.}
	\label{tab:tab1}
    \begin{tabular}{c|c|c}
        \toprule[1.2pt]
        \multirow{2}{*}{Method} & \multicolumn{1}{c|}{PTIS dataset} & \multicolumn{1}{c}{RISMR dataset}\\
        & SIQS $\uparrow$ & SIQS$\uparrow$\\
        \midrule \midrule
        SPW~\cite{SPW} & 7.26 & 7.11\\
        LPC~\cite{LPC} & 7.49 & 7.64 \\
        \midrule
        UDIS++~\cite{UDIS++} & 7.63 & 6.69 \\
        SRStitcher~\cite{SRStitcher} & 8.00 & 7.88\\
        PIS3R/LiftProj-style$^\dagger$ & \multirow{2}{*}{7.91} & \multirow{2}{*}{8.36}\\
        \cite{PIS3R,LiftProj} & & \\
        GPIS~\cite{GPIS} & 6.06 & --\\
        RopStitch~\cite{RopStitch} & 5.57 & 4.57\\
        \midrule
        AnyStitch & \textbf{9.40} & \textbf{9.71} \\
        \bottomrule[1.2pt]
    \end{tabular}
\end{table}

\paragraph{Comparison methods and benchmarks}
Since many recent advancements in image stitching focus on two-view inputs, we first evaluate our approach by comparing it with existing two-view methods, including traditional methods SPW~\cite{SPW}, LPC~\cite{LPC} and learning-based methods UDIS++~\cite{UDIS++}, SRStitcher \cite{SRStitcher}, PIS3R~\cite{PIS3R}, LiftProj~\cite{LiftProj}, GPIS~\cite{GPIS} and RopStitch~\cite{RopStitch}. Notably, similar to our approach, SRStitcher and GPIS also leverage generative models for image stitching, while PIS3R and LiftProj employ 3D-based models to establish geometry-aware correspondences under large parallax, making them highly relevant competitors for evaluating the superiority of our method.
Since the official implementations of PIS3R and LiftProj were unavailable, we reproduce their shared reconstruction-and-projection pipeline by estimating a point cloud with VGGT, projecting it to the target view, and applying a generative restoration model fine-tuned on AnyStitchSyn. We denote this reproduction as PIS3R/LiftProj-style$^\dagger$ rather than reporting it as an official result. 
We conduct experiments on our AnyStitchSyn test set and two challenging public benchmarks: the PTIS dataset~\cite{PTIS} and the RISMR dataset~\cite{RISMR}.
These datasets are characterized by scenes exhibiting large parallax, making them ideal testing cases for assessing the robustness of different image stitching algorithms.
Additionally, we compare our method against the state-of-the-art multi-view image stitching method~\cite{GES-GSP} on standard classical scenes and our custom large-parallax scenes, evaluating the robustness of AnyStitch with multiple inputs.

\begin{figure}[!t]
	\centering
	\includegraphics[width=\linewidth]{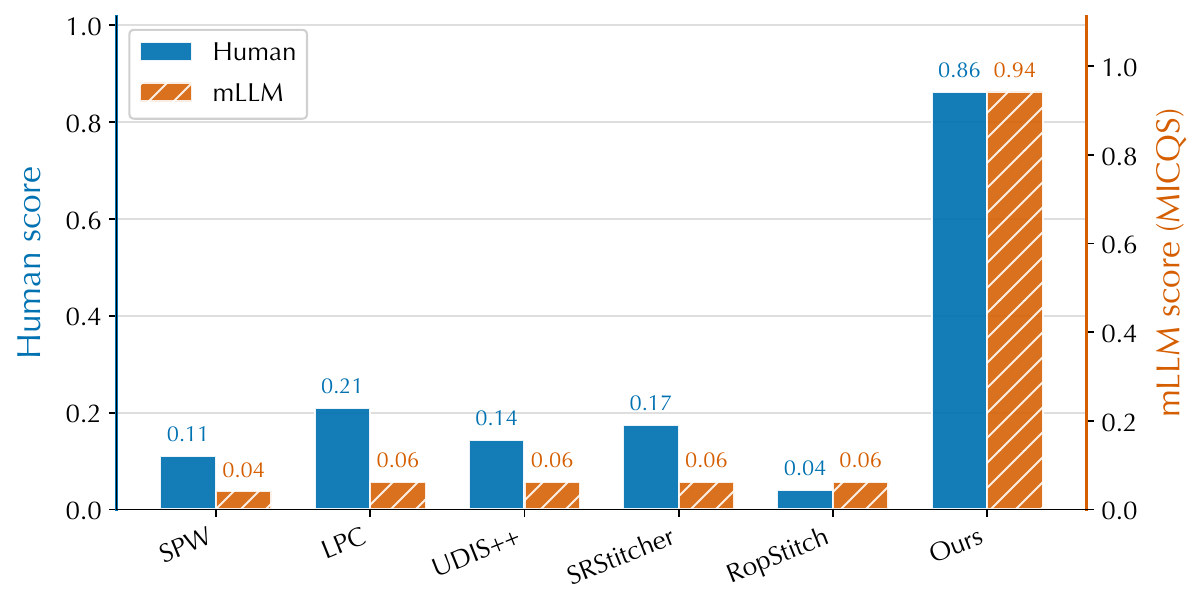}
	\caption{User study and mLLM evaluation on two large-parallax datasets.}
    \label{fig:user_study}
\end{figure}

\begin{figure*}[!t]
	\centering
	\includegraphics[width=\linewidth]{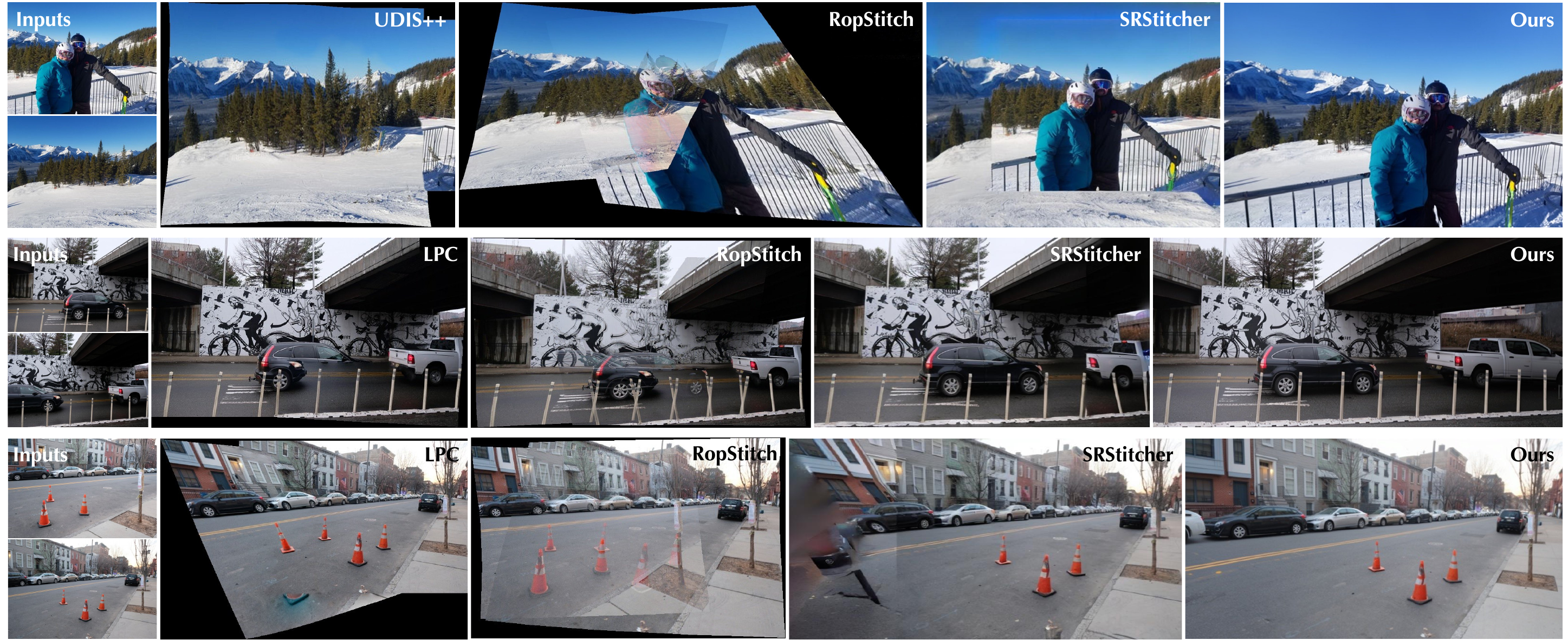}
	\caption{Visual comparisons on RISMR~\cite{RISMR} dataset.}
    \label{fig:comparison_riswmr}
\end{figure*}

\begin{figure*}[!t]
	\centering
	\includegraphics[width=\linewidth]{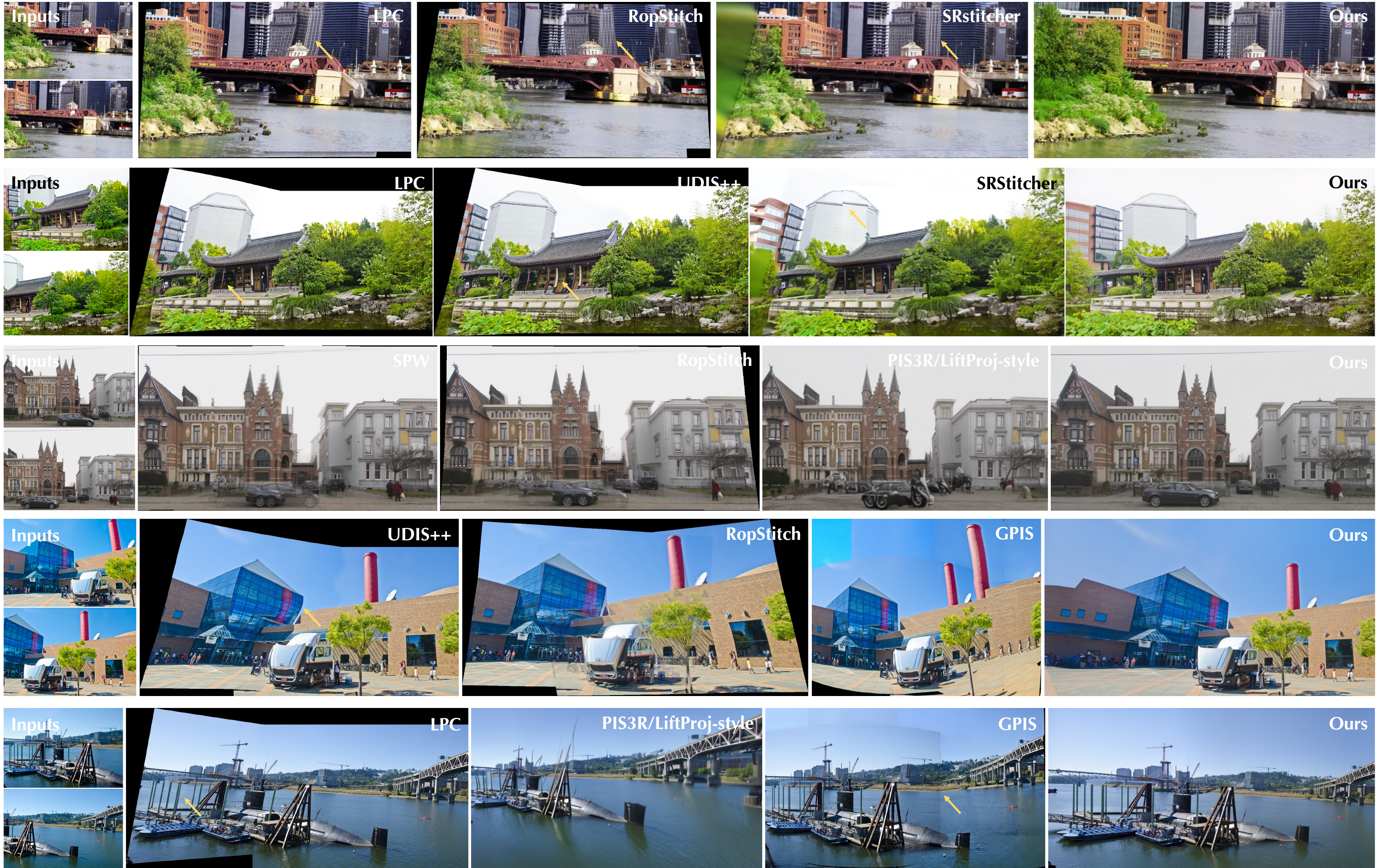}
	\caption{Visual comparisons on the PTIS dataset~\cite{PTIS}.}
    \label{fig:comparison_ptis}
\end{figure*}

\subsection{Comparison results}
\paragraph{Two-view comparison}
{
We first evaluate on the AnyStitchSyn test set, which provides pixel-aligned ground-truth panoramas, using full-reference metrics SSIM and LPIPS. We compare AnyStitch with UDIS++, SRStitcher, and RopStitch. For a fair evaluation, AnyStitch estimates the input depth and camera poses using Depth Anything 3 (DA3)~\cite{depthanything3}, rather than using the rendered ground-truth geometry available in the synthetic dataset. To prevent irregular black boundaries from biasing the metrics, for each method, we compute the metrics over the non-black region shared by its output and the ground truth. As reported in Tab.~\ref{tab:full_reference}, AnyStitch achieves the best result on both metrics, indicating that our method better handles large parallax while preserving content fidelity. The qualitative comparisons in Fig.~\ref{fig:comparison_anystitchsyn} further show that AnyStitch produces fewer structural distortions and better preserves the scene content relative to the ground truth.
}

We then perform a reference-free evaluation on the two real-world large-parallax benchmarks.
We summarize the quantitative comparison results in Tab.~\ref{tab:tab1}. Our AnyStitch achieves the best performance, indicating that our method outputs the stitched panorama with superior visual quality and the most plausible semantic coherence. Specifically, our method outperforms the generative method SRStitcher in a large margin on both datasets in terms of SIQS score (9.40/9.71 \textit{v.s.} 8.00/7.88). Beyond individual image quality assessment, Fig.~\ref{fig:user_study} compares our method with the baselines through human and mLLM-based MICQS evaluations. These results are consistent with the quantitative findings in Tab.~\ref{tab:tab1}.

\begin{figure*}[!t]
	\centering
	\includegraphics[width=\linewidth]{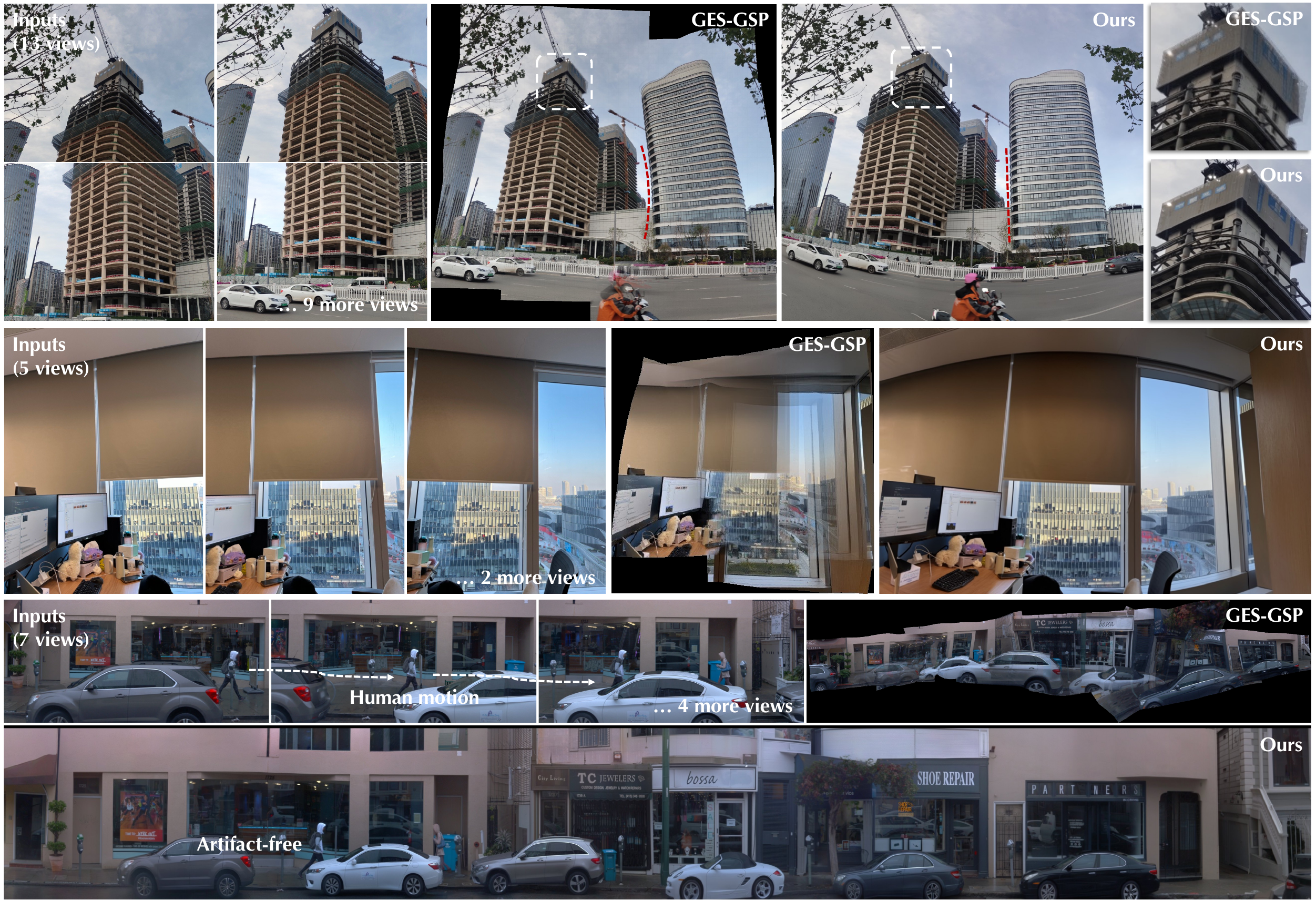}
	\caption{Comparison on multi-view scenes. Top and middle scenes: equirectangular projection. Bottom scene: parallel projection.}
    \label{fig:comparison_multiview}
\end{figure*}

The qualitative comparison results are visualized in Fig.~\ref{fig:comparison_riswmr} and Fig.~\ref{fig:comparison_ptis}.
Traditional and learning-based frameworks, such as LPC, UDIS++, and RopStitch, struggle with depth disparities. Because they rely on 2D homography or localized 2D deformation fields, and often forcibly warp non-planar regions to satisfy feature alignment, resulting in unnatural spatial stretching and severe geometric distortions, as evident on buildings and the temple pillar in Fig~\ref{fig:comparison_ptis}. Furthermore, as they fail to disentangle complex 3D foreground-background relationships, these methods introduce severe ghosting artifacts on distinct foreground objects, such as people, moving cars and traffic cones in Fig~\ref{fig:comparison_riswmr}.
While SRStitcher and GPIS employ generative models to facilitate stitching and hallucinate missing regions, they lack explicit and precise 3D geometric guidance. Consequently, their results frequently exhibit structural distortions, visible seams, and blurry content, as we marked in Fig.~\ref{fig:comparison_ptis}. Although the PIS3R/LiftProj-style baseline uses explicit 3D geometry to guide stitching, its pixel-level projection can lead to severe content distortions.

In contrast, AnyStitch generates visually stunning, artifact-free panoramas that adhere to the real-world's structure. By grounding DiT with the precise hierarchical projected RoPE derived from explicit 3D geometry, our method perfectly aligns intricate architectural lines (see Fig.~\ref{fig:comparison_ptis}) without structural degradation. Moreover, AnyStitch can correctly handle the large parallax between foreground and background, rendering artifact-free panoramas (see Fig.~\ref{fig:comparison_riswmr}).

\paragraph{Multi-view comparison}
Fig.~\ref{fig:comparison_multiview} presents a qualitative comparison between AnyStitch and the state-of-the-art multi-view method GES-GSP~\cite{GES-GSP} on three scenarios with increasing camera translation and parallax. In the top scene, which is a simple case from nearly without camera translation. However, GES-GSP still suffers from overlay artifacts and structure distortion. In the middle and bottom scene, where severe parallax occurs, GES-GSP degrades remarkably. In contrast, our AnyStitch is robust to both tiny-parallax and large-parallax scenes, producing visually pleasing panoramas. Moreover, as shown in the bottom case, AnyStitch effectively handles simple isolated human motion, enabled by the full attention mechanism of the DiT architecture.

\begin{figure*}[!t]
	\centering
	\includegraphics[width=\linewidth]{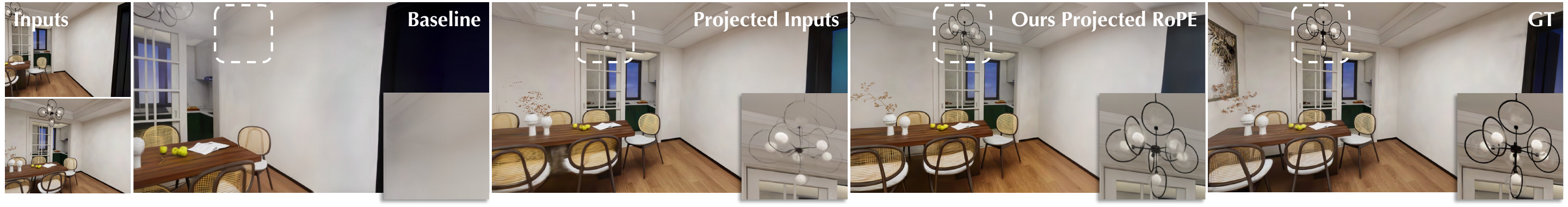}
	\caption{Visual comparisons of different geometric injection methods. Our method uses GT geometry.}
    \label{fig:ab_rope_syn}
\end{figure*}

\begin{figure*}[!t]
	\centering
	\includegraphics[width=\linewidth]{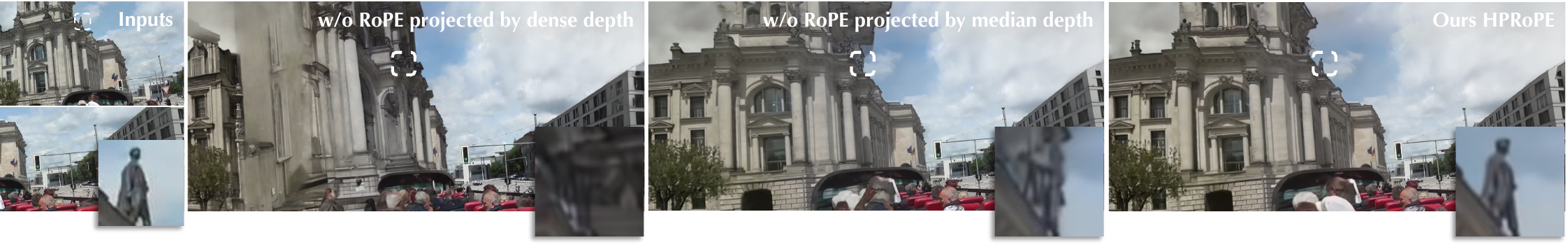}
	\caption{Visual effectiveness of using two complementary projected RoPE.}
    \label{fig:ab_rope}
\end{figure*}

\begin{table}[!t]
    \centering
    \caption{Quantitative ablation study on HPRoPE. The geometry-conditioned variants use the ground-truth camera poses and rendered depth maps from the 3DGS scenes.}
	\label{tab:ablation}
    \begin{tabular}{c|cc}
        \toprule[1.2pt]
        \multirow{2}{*}{Model} & \multicolumn{2}{c}{AnyStitchSyn$^{\mathrm{w/\ GT\ geometry}}$} \\
        & PSNR$\uparrow$ & SSIM$\uparrow$ \\
        \midrule 
        \midrule
        baseline & 13.71 & 0.7646 \\
        baseline + projected inputs & 20.72 & 0.8744 \\
        (Ours) baseline + HPRoPE & \textbf{21.39} & \textbf{0.8806}\\
        \bottomrule[1.2pt]
    \end{tabular}
\end{table}

\subsection{Ablation Study}
To validate the effectiveness of our proposed architectural designs, we conduct comprehensive ablation studies on two key components, \textit{i.e}., Hierarchical Projected RoPE (HPRoPE) and our autoregressive generation strategy. 

\paragraph{Hierarchical Projected RoPE}
The accurate geometric perception capability of our model relies on the injection of spatial correspondence. To investigate this, we compare three variants. The first variant (denoted as \textit{baseline}) is the vanilla DiT, which is completely devoid of any explicit geometric priors. The second variant (denoted as \textit{baseline + projected inputs}) utilizes estimated 3D information to project raw inputs to the panoramic canvas before tokenization. This variant still employs standard 2D positional encodings. The final variant is our proposed HPRoPE (denoted as \textit{baseline + HPRoPE}), which directly modifies the positional encodings to inject precise projection coordinates into the latent space.
We conduct these evaluations on the test set of our AnyStitchSyn Dataset, measuring the performance using PSNR and SSIM. Unlike the full-reference comparison in Tab.~\ref{tab:full_reference}, the geometry-conditioned variants in this controlled ablation use the ground-truth camera poses and rendered depth maps from the 3DGS scenes. The quantitative and qualitative results are presented in Tab.~\ref{tab:ablation} and Fig.~\ref{fig:ab_rope_syn}, respectively. 

As shown, the baseline variant entirely depends on its inherent self-attention mechanism to infer spatial correspondences across the unorganized, large-parallax input views. As shown in Tab.~\ref{tab:ablation} and Fig.~\ref{fig:ab_rope_syn}, this unconstrained generation struggles significantly, yielding poor performance (13.71 PSNR and 0.7646 SSIM). 
The second variant attempts to mitigate this by introducing explicit spatial modeling in the image space.  Although it forces the model to synthesize a structurally correct panorama (significantly boosting PSNR to 20.72), pixel-imperfect depth will cause the content distortion in the projected inputs. Consequently, it suffers from noticeable information loss and drift, as indicated by Fig.~\ref{fig:ab_rope_syn}.
Instead, our proposed method resolves these issues by explicitly feeding the geometric mapping into the attention mechanism via Hierarchical Projected RoPE. As a result, our method achieves the highest quantitative performance (21.39 PSNR and 0.8806 SSIM). Fig.~\ref{fig:ab_rope_syn} also demonstrates that our approach can effectively avoid content loss and distortion compared to performing projection in image space.

\paragraph{In-depth analysis of HPRoPE}
We further investigate the impact of the introduction of two distinct depth projection, \textit{i.e.}, accurate dense depth and the median depth. As illustrated in Fig.~\ref{fig:ab_rope}, relying exclusively on either single depth projection produces suboptimal results.
On the one hand, without RoPE projected by dense depth, it fails to generate results with accurate 3D geometry, leading to severe content change. 
On the other hand, using RoPE projected by dense depth as the sole geometric guidance may also cause the content change due to the inaccurate depth estimation, especially in the transition regions between the foreground and the background. As shown in the zoomed-in region of Fig.~\ref{fig:ab_rope}, the statue on the roof is severely distorted.
Our proposed hierarchical projected RoPE resolves this dilemma by simultaneously feeding both geometric cues into disjoint subsets of attention heads. This dual-precision mechanism balances geometrical alignment and the fidelity preservation, ensuring the generation of a complete statue in Fig~\ref{fig:ab_rope} under the view change.
We provide an additional detailed analysis under imperfect geometry in supplementary material.

Next, we evaluate the effectiveness of our patch-wise autoregressive generation mechanism for flexible panoramas generation.

\begin{figure}[!t]
	\centering
	\includegraphics[width=\linewidth]{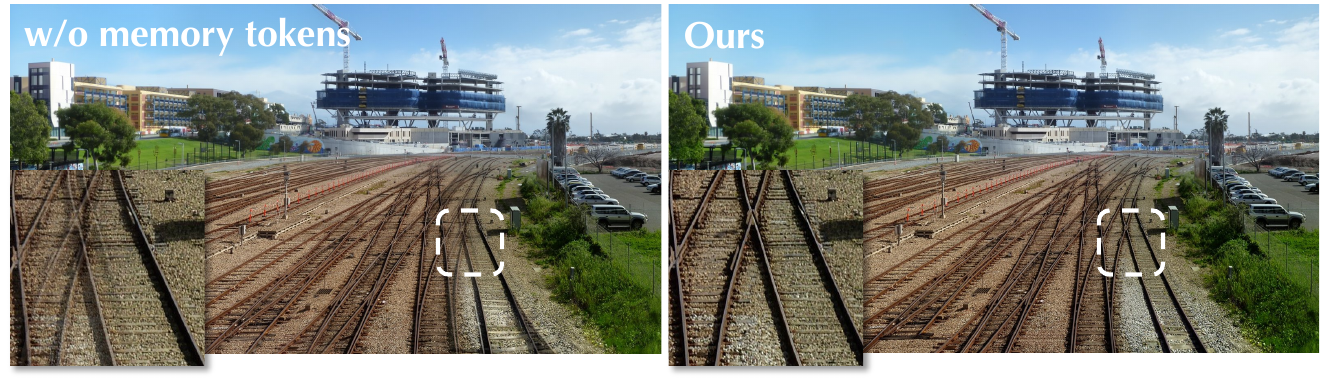}
	\caption{Visual effectiveness of introducing memory tokens.}
    \label{fig:ab_ar_consistency}
\end{figure}

\paragraph{Memory tokens}
We first investigate the critical role of memory tokens during the progressive patch generation process. As shown in Fig.~\ref{fig:ab_ar_consistency}, generating adjacent spatial patches without the injection of memory tokens leads to the overlay artifacts. This failure occurs because, in the absence of memory tokens, the DiT processes each local generation window independently, lacking the contextual awareness of the previously generated regions.
By concatenating memory tokens sampled from the overlapping areas, our method seamlessly anchors the newly generated latent features to the existing global canvas, ensuring the overall consistency when generating a large panorama.

\begin{figure}[!t]
	\centering
	\includegraphics[width=\linewidth]{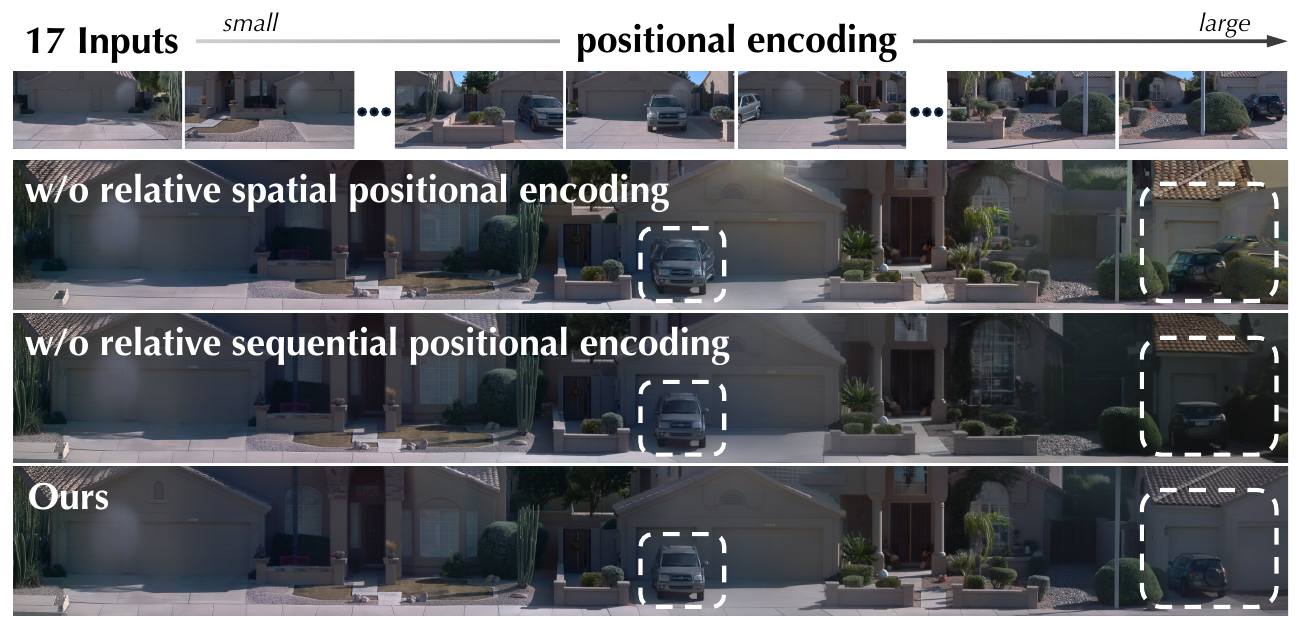}
	\caption{Visual effectiveness of relative positional encoding.}
    \label{fig:ab_ar_relative_rope}
\end{figure}

\paragraph{Relative positional encoding}
Second, we investigate the necessity of the relative positional encoding when processing a large number of views (\textit{e.g.}, 17 inputs captured in a long street, as shown in Fig.~\ref{fig:ab_ar_relative_rope}). In such scenarios, absolute positional coordinates inevitably grow beyond the distribution observed during DiT training, triggering extrapolation collapse. When the relative spatial normalization is disabled (\textit{w/o relative spatial positional encoding}), the spatial frequencies shift out of the stable pre-trained range. As highlighted in the rightmost dashed box, this spatial extrapolation failure causes the network to generate noticeable artifacts. Similarly, disabling the relative normalization for the input sequence index (\textit{w/o relative sequential positional encoding}) leads to sequential extrapolation failure.
By integrating these components, we constrain all positional encodings within a stable, trained range. As demonstrated in the bottom row of Fig.~\ref{fig:ab_ar_relative_rope}, our AnyStitch is able to process a variable number of inputs and output seamless contents across a large canvas.

\begin{figure*}[!t]
	\centering
	\includegraphics[width=\linewidth]{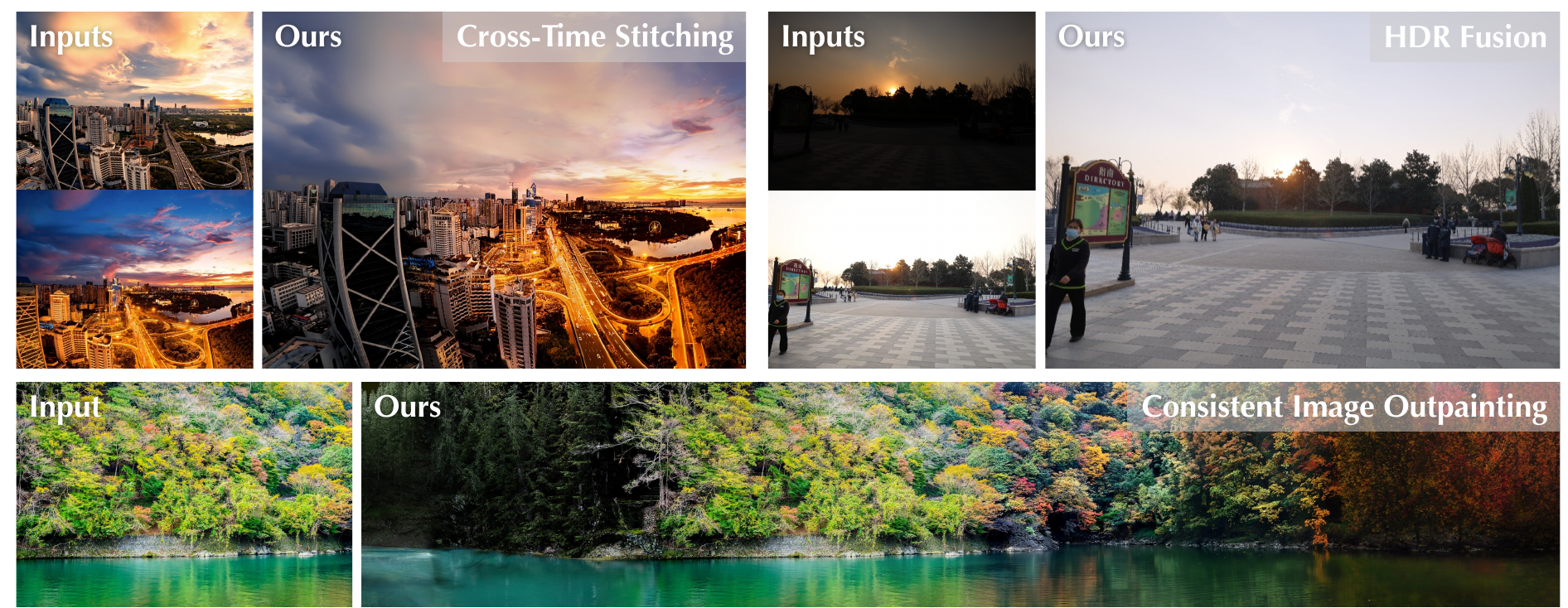}
	\caption{More applications of our AnyStitch.}
    \label{fig:application}
\end{figure*}

\paragraph{Window size and stride}
Last, we evaluate four generation-window sizes while fixing the overlap ratio at 25\%. All timings reported in Tab.~\ref{tab:window_stride} use 40 denoising steps and are measured on panoramas from AnyStitchSyn test set with an average resolution of $3337\times1586$. Although the $1536/1152$ setting yields the highest PSNR and the $2048/1536$ setting yields the highest SSIM, the default $1024/768$ setting provides the best overall trade-off between reconstruction quality and inference time.

\begin{table}[!t]
    
    \centering
    \caption{Ablation of generation-window and stride sizes. Runtime is measured on panoramas with an average resolution of $3337\times1586$.}
    \label{tab:window_stride}
    \begin{tabular}{c|c|c|c}
        \toprule[1.2pt]
        \multirow{2}{*}{Window/stride} & \multicolumn{3}{c}{AnyStitchSyn$^{\mathrm{w/\ GT\ geometry}}$} \\
        & PSNR $\uparrow$ & SSIM $\uparrow$ & Time $\downarrow$ \\
        \midrule \midrule
        512/384 & 20.13 & 0.8597 & 7.5 min \\
        1024/768 & 21.39 & 0.8806 & \textbf{7.2 min} \\
        1536/1152 & \textbf{21.59} & 0.8795 & 13.6 min \\
        2048/1536 & 21.33 & \textbf{0.8816} & 10.6 min \\
        \bottomrule[1.2pt]
    \end{tabular}
\end{table}

\subsection{More Applications}
As a generative stitching framework, AnyStitch can move far beyond mere image stitching. As shown in Fig.~\ref{fig:application}, it can generalize to cross-time stitching, HDR fusion of misaligned exposures, and consistent image outpainting, further highlighting the robustness and potential of our method.

\section{Limitations}
\begin{figure}[!t]
	\centering
	\includegraphics[width=\linewidth]{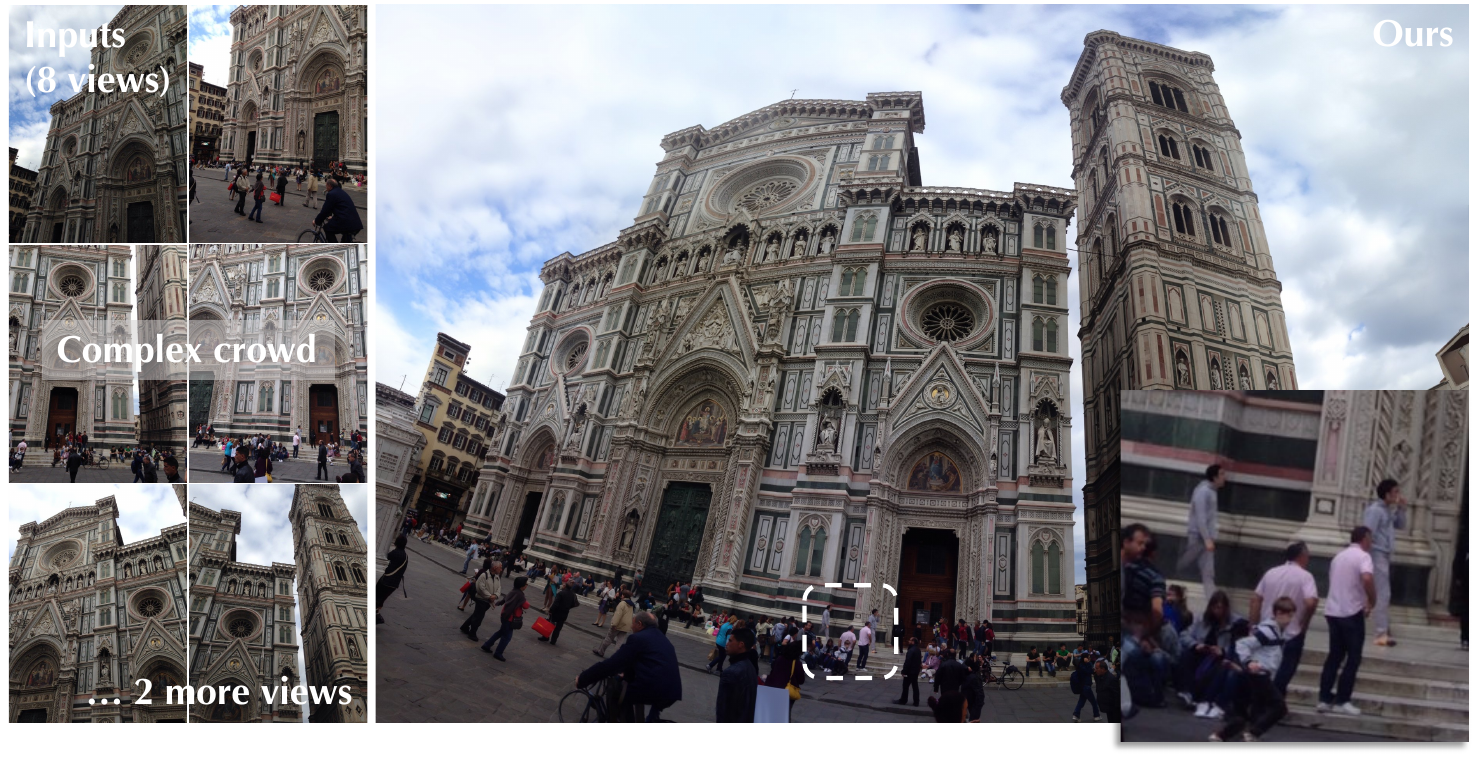}
	\caption{A failure case in a crowded dynamic scene. Given eight input views captured at different times, moving pedestrians may be duplicated or appear ghosted in the stitched panorama, as highlighted in the zoomed-in region.}
	\label{fig:failure_case}
\end{figure}

Although our experiments demonstrate AnyStitch's strong potential for casual-photography panorama stitching, several limitations remain. First, applying AnyStitch to measurement-critical applications, such as mapping or satellite imagery, may pose some fidelity risks due to its generative nature. Second, as a 40-step diffusion model, AnyStitch is slower than conventional warping-based methods: generating a 2K panorama takes approximately 5 minutes and requires about 40~GB of GPU memory on one NVIDIA A800 equivalent GPU. Finally, although the method exhibits some tolerance to isolated moving subjects, complex crowded scenes may still produce motion-related duplication and ghosting artifacts, as shown in Fig.~\ref{fig:failure_case}. Reducing the number of sampling steps through distillation and incorporating dynamic-scene training data are promising directions for future work.

\section{Conclusion}
In this paper, we present AnyStitch, a novel generative framework that tackles the highly challenging problem of large-parallax image stitching. By reformulating the stitching pipeline from traditional 2D pixel-space warping to a reference-based generative process, we effectively circumvent the persistent ghosting, seams, and structural tearing. To overcome the spatial misalignment issues inherent in standard diffusion models, we introduce the Hierarchical Projected Rotary Positional Encoding (HPRoPE). This mechanism injects 3D geometric priors directly into the latent space, utilizing varying attention-head precisions to balance rigorous structural constraints with high-fidelity content generation. Furthermore, we design a patch-wise spatial autoregressive strategy that bypasses the computational resolution limits of generative models. To train our model, we synthesize large-parallax inputs with corresponding ground truths from 3DGS data. With these features, AnyStitch enables the seamless integration of a variable number of casual captures into visually coherent panoramas. Experiments demonstrate that AnyStitch consistently outperforms both existing two-view methods and multi-view frameworks.

\begin{acks}
This work was supported by the National Key R\&D Program of China (No.~2025YFE0201300) and the Research Grants Council of Hong Kong under the Early Career Scheme (ECS) (No.~24209224). Additional support was provided by the CUHK-CUHK(SZ)-GDSTC Joint Collaboration Fund (No.~2025A0505000053). The work was also supported in part by the Centre for Perceptual and Interactive Intelligence (CPII) Ltd., a CUHK-led InnoCentre under the InnoHK initiative of the Innovation and Technology Commission of the Hong Kong Special Administrative Region Government.
\end{acks}

\bibliographystyle{ACM-Reference-Format}
\bibliography{references}

\clearpage
\appendix
\section{Training Details}
To train AnyStitch, we establish a training paradigm designed to bridge the train-test gap. Initially, we derive the HPRoPE directly utilizing the rendered depth and camera poses. To replicate the spatial autoregressive behavior of the testing phase, we implement an overlap-aware sampling strategy during training. Specifically, we randomly extract tokens from ground truth to serve as memory tokens, which are then concatenated with input tokens. More importantly, we propose a novel sequence ID augmentation operation that assigns random sequence indices to inputs during training. This strategy disentangles the generative priors from specific input lengths, enabling our AnyStitch to be efficiently trained on only two-view pairs while generalizing to multi-view inputs.

\section{More Ablations}

\subsection{Sequential ID Augmentation}

\begin{figure*}[!t]
	\centering
	\includegraphics[width=\linewidth]{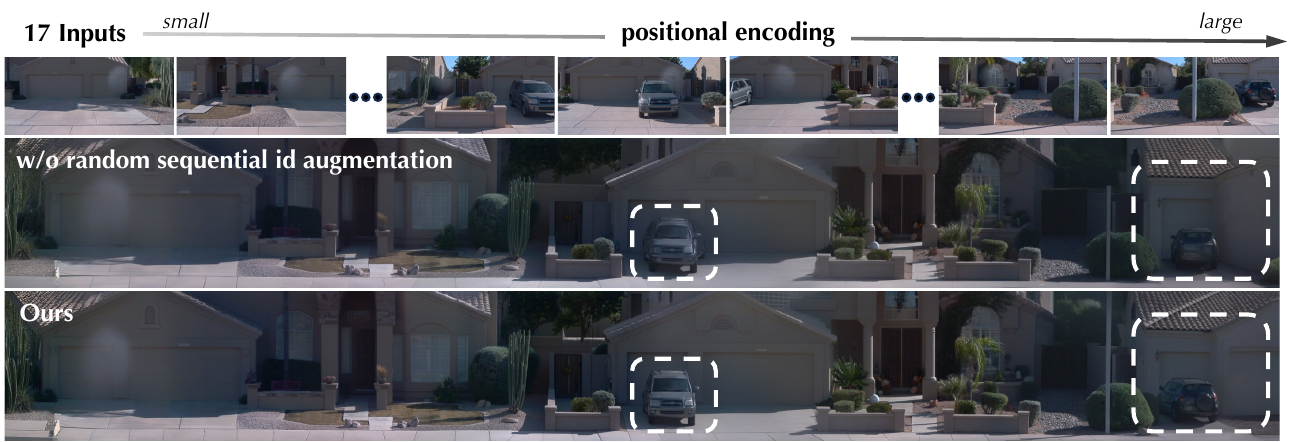}
	\caption{Visual effectiveness of sequential id augmentation strategy.}
    \label{fig:id_aug}
\end{figure*}

To validate the effectiveness of this augmentation strategy, we train another variant denoted as \textit{w/o random sequential id augmentation}. As shown in Fig.~\ref{fig:id_aug}, without this dynamic augmentation during training, the network tends to overfit a fixed, limited number of inputs (we only use 2 inputs in our training data) and struggles to generalize to long sequences like 17 inputs, again yielding distorted details.

\subsection{Projection-Accuracy Ablation}
\label{sec:supp_projection_accuracy}

\begin{figure*}[!t]
    \centering
    \includegraphics[width=\linewidth]{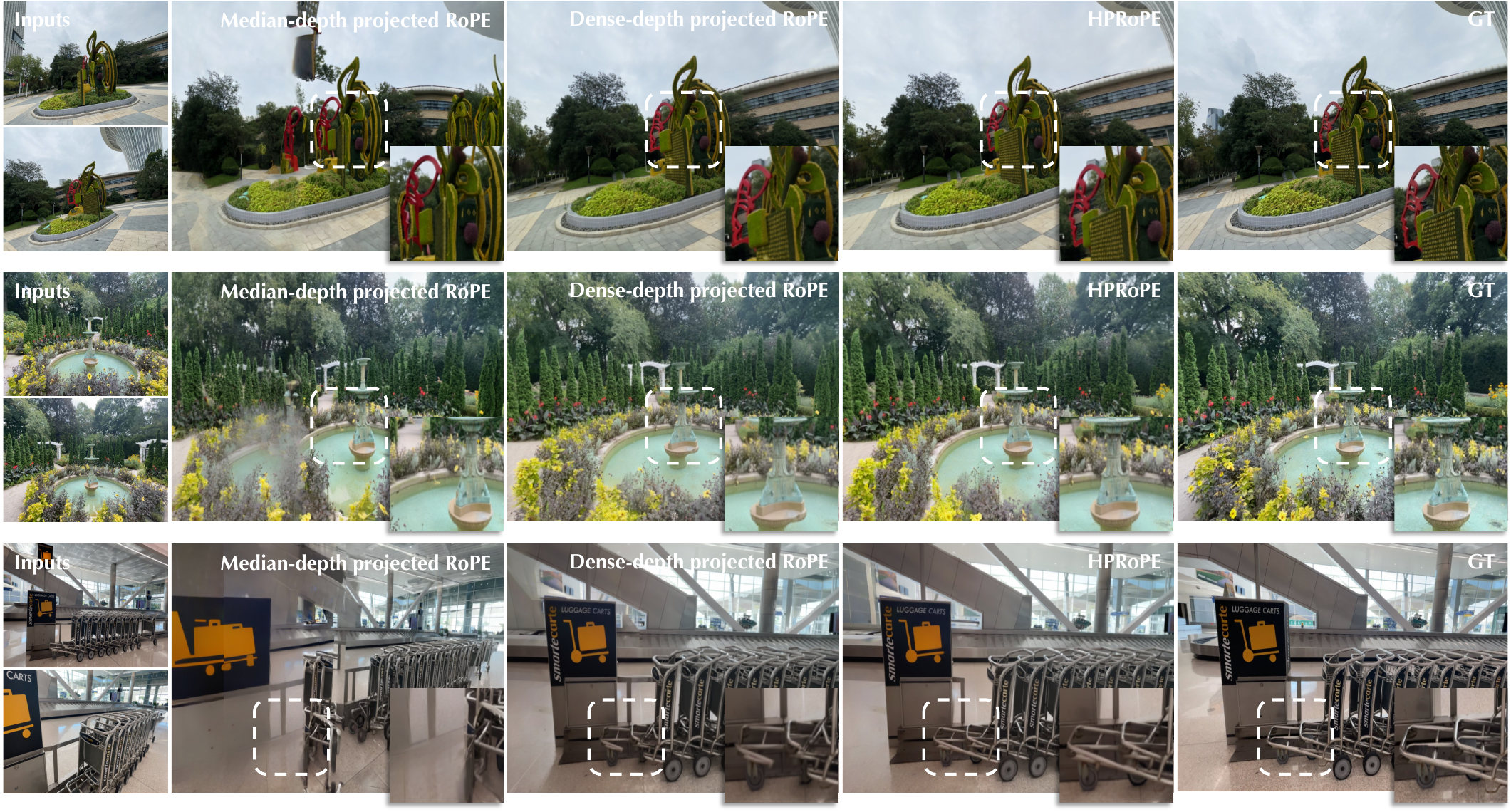}
    \caption{Qualitative projection-accuracy ablation on real-world DL3DV-10K scenes using DA3-estimated camera poses and depth.}
    \label{fig:ab_dl3dv}
\end{figure*}

To further quantitatively evaluate the effectiveness of our dual-precision RoPE under imperfect geometry, we conduct an additional experiment on real-world scenes from the DL3DV-10K benchmark~\cite{DL3DV}, using camera poses and depth estimated by DA3~\cite{depthanything3}. For each scene, we sample three temporally ordered frames, using the first and last frames as input views and the middle frame to define the target viewpoint and serve as ground truth. We generate a stitched result at this target viewpoint from the two input views and compare it against the middle frame. We adopt the shift-tolerant perceptual metric DISTS~\cite{DISTS}, where a lower score indicates better agreement with the reference view. As shown in Tab.~\ref{tab:projection_accuracy}, dense-depth projection substantially improves upon median-depth projection, while HPRoPE achieves the lowest DISTS by combining both geometric cues. The qualitative comparisons in Fig.~\ref{fig:ab_dl3dv} further show that HPRoPE better preserves local structure and content fidelity.

\begin{table}[h]
    
    \centering
    \caption{Projection-accuracy ablation on real-world DL3DV-10K benchmark using DA3-estimated camera poses and depth.}
    \label{tab:projection_accuracy}
    \begin{tabular}{c|c}
        \toprule[1.2pt]
        Projection strategy & DISTS $\downarrow$ \\
        \midrule \midrule
        Median-depth projection & 0.1872 \\
        Dense-depth projection & 0.0967 \\
        HPRoPE & \textbf{0.0902} \\
        \bottomrule[1.2pt]
    \end{tabular}
\end{table}

\subsection{Mixed-Projection Training Ablation}
\label{sec:supp_mixed_projection}

\begin{figure*}[!t]
    \centering
    \includegraphics[width=\linewidth]{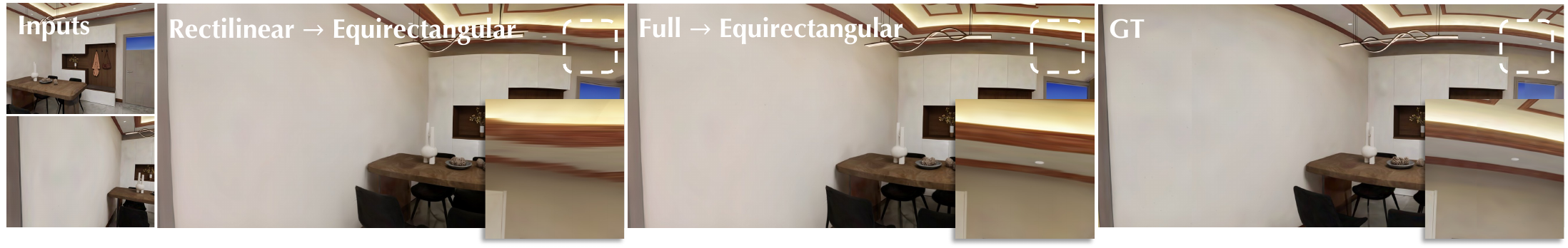}
    \caption{Qualitative cross-projection ablation on the AnyStitchSyn test set.}
    \label{fig:ab_projection_type}
\end{figure*}

We compare projection-specific variants trained using only rectilinear or equirectangular targets with our full mixed-projection training on the AnyStitchSyn test set. As shown in Tab.~\ref{tab:mixed_projection}, the rectilinear-only model obtains 21.30 dB when evaluated with equirectangular projection, whereas mixed-projection training improves the result to 22.08 dB. In the reverse setting, the equirectangular-only model obtains 21.10 dB on rectilinear targets, while mixed-projection training reaches 21.39 dB. These results indicate that exposure to diverse projection geometries helps the model adapt to the intra-token deformations induced by different projection types. The qualitative comparison in Fig.~\ref{fig:ab_projection_type} further shows that train-test gap may cause artifacts due to unseen token deformation. Training with mixed projection types can effectively mitigate this phenomenon.

\begin{table}[h]
    
    \centering
    \caption{Cross-projection ablation on the AnyStitchSyn test set. Mixed-projection training uses the full set of projection types.}
    \label{tab:mixed_projection}
    \begin{tabular}{c|c|c}
        \toprule[1.2pt]
        Training setting & Test projection & PSNR $\uparrow$ \\
        \midrule \midrule
        Rectilinear-only & Equirectangular & 21.30 \\
        Mixed projections & Equirectangular & \textbf{22.08} \\
        \midrule
        Equirectangular-only & Rectilinear & 21.10 \\
        Mixed projections & Rectilinear & \textbf{21.39} \\
        \bottomrule[1.2pt]
    \end{tabular}
\end{table}

\section{Projected-Input Coverage Visualization}
\label{sec:supp_coverage}

\begin{figure*}[!t]
    \centering
    \includegraphics[width=\linewidth]{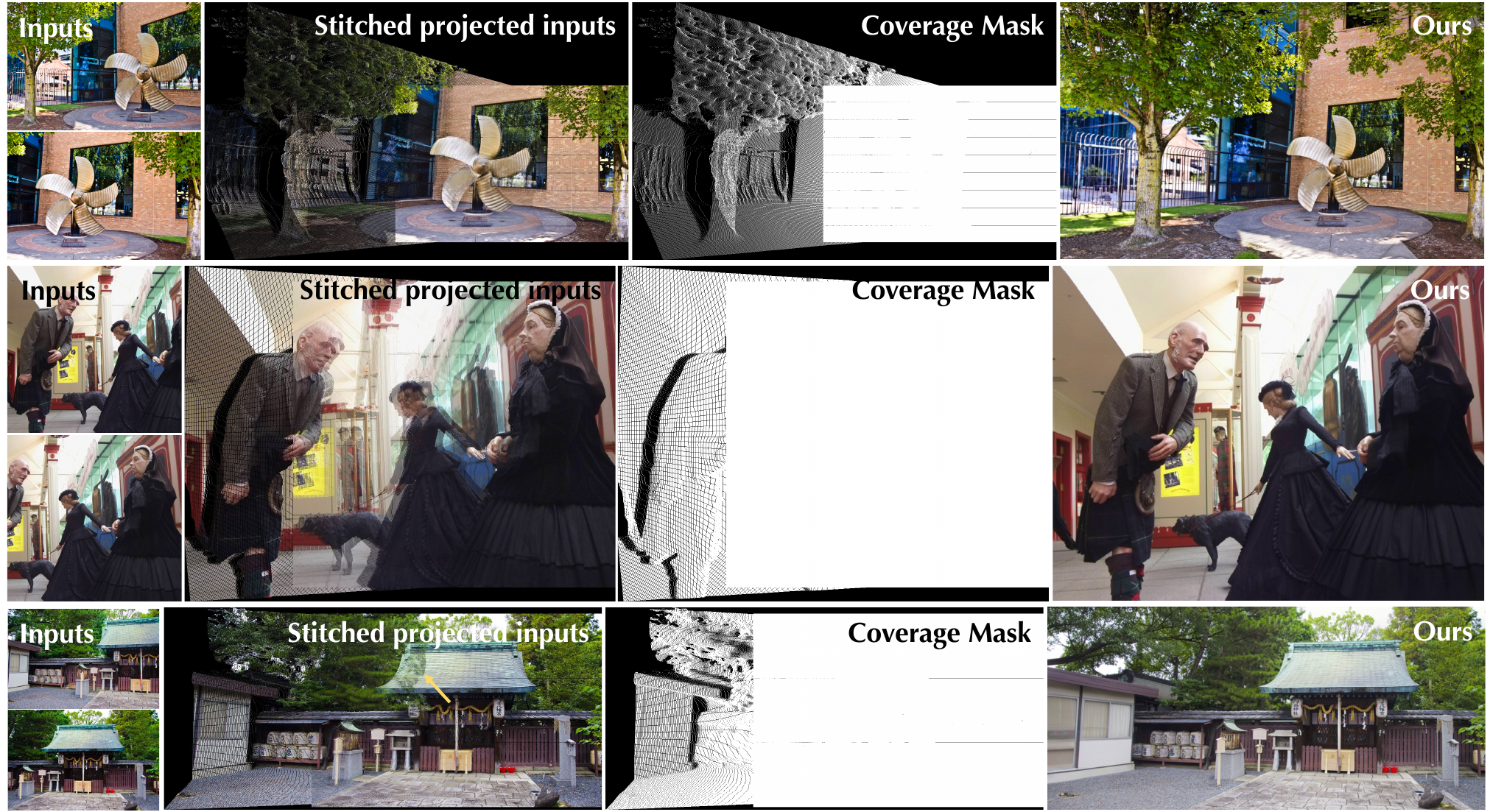}
    \caption{Projected-input coverage visualization. From left to right: input views, RGB inputs projected onto the target canvas for visualization, projected-input coverage mask, and the AnyStitch result. Bright mask regions have projected input-token support, while dark regions lack such support and are completed by the generative model.}
    \label{fig:coverage_mask}
\end{figure*}

Although AnyStitch does not warp the input RGB content during generation, we project the inputs onto the target canvas here solely for visualization. We obtain the projected-input coverage mask by rasterizing the projected token coordinates used by HPRoPE. Bright mask regions indicate target-canvas locations supported by at least one projected input token, whereas dark regions have no projected input support and require generative completion. As shown in Fig.~\ref{fig:coverage_mask}, AnyStitch can handle occlusions (top row), inaccurate projections (middle row) and depth-ordering conflicts (bottom row), thus producing coherent panoramas across both supported and uncovered regions.

\begin{figure*}[!t]
	\centering
	\includegraphics[width=\linewidth]{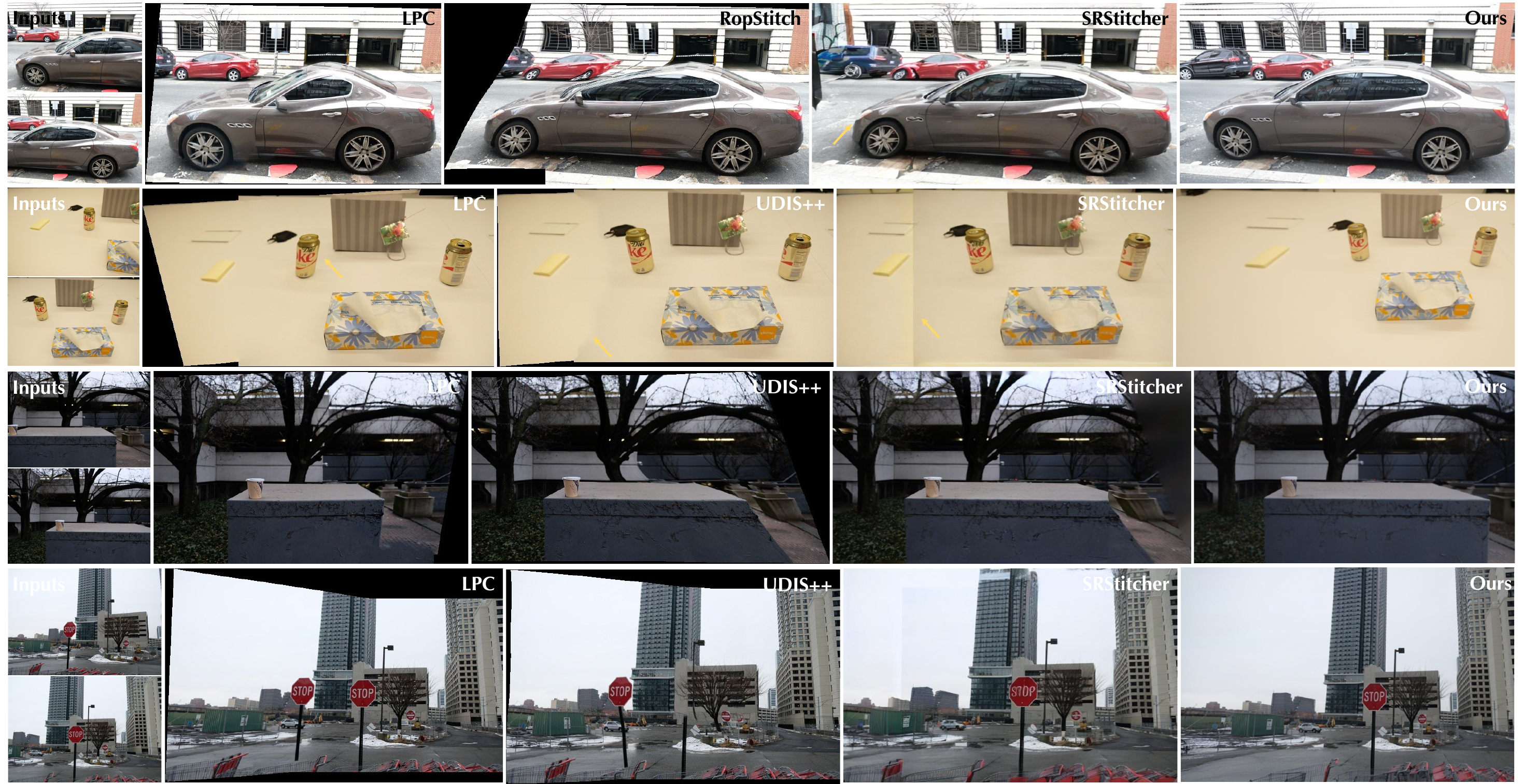}
	\caption{More visual results on RISMR dataset~\cite{RISMR}.}
    \label{fig:comparison_riswmr_supp}
\end{figure*}

\begin{figure*}[!t]
	\centering
	\includegraphics[width=\linewidth]{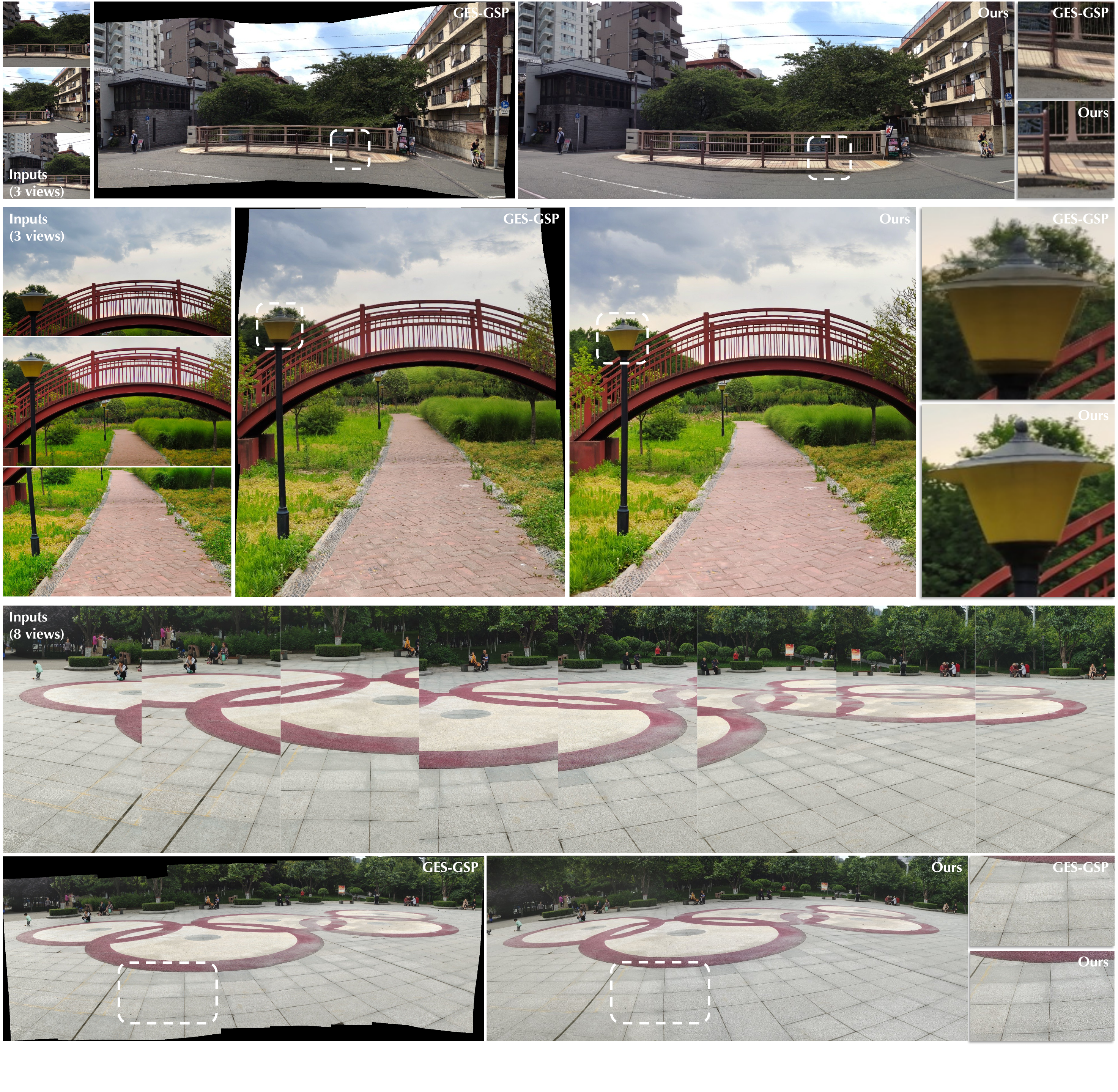}
	\caption{More visual results on classical stitching scenes.}
    \label{fig:comparison_multiview_supp}
\end{figure*}

\section{More Visual Results}

To further validate the superior performance of our proposed method, we provide more visual comparisons with two-view methods and multi-view methods in Fig.~\ref{fig:comparison_riswmr_supp} and Fig.~\ref{fig:comparison_multiview_supp}, respectively.

\end{document}